\documentclass[lettersize,journal]{IEEEtran}

\usepackage{amsmath,amsfonts}
\usepackage{array}
\usepackage{graphicx}
\usepackage{cite}
\usepackage{url}
\usepackage{textcomp}
\usepackage{stfloats}
\usepackage{verbatim}

\usepackage[caption=false,font=normalsize,labelfont=sf,textfont=sf]{subfig}

\usepackage{amssymb}
\usepackage{bm}
\usepackage{booktabs}
\usepackage{makecell}
\usepackage{afterpage}

\usepackage[ruled,linesnumbered,noend]{algorithm2e}

\usepackage[hidelinks]{hyperref}

\newcommand{\figcaption}[1]{...}
\newcommand{\tblcaption}[1]{...}

\newcolumntype{C}[1]{...}

\newcommand{\figureN}{Fig.~}
\newcommand{\tableN}{TABLE~}
\def\fnum@table{\tableN\thetable}

\newcommand{\equationname}[1]{Eq.~(#1)}

\begin{document}

\title{
DecompGrind: A Decomposition Framework for Robotic Grinding via Cutting-Surface Planning and Contact-Force Adaptation
}

\author{Shunsuke Araki$^{1}$, Takumi Hachimine$^{1}$, Yuki Saito$^{2}$, Kouhei Ohnishi$^{2}$, Jun Morimoto$^{3,4}$, \\ and Takamitsu Matsubara$^{1}$
\thanks{This work was supported by JST-Mirai Program Grant Number JPMJMI21B1, Japan.}
\thanks{$^{1}$S. Araki, T. Hachimine~and~T. Matsubara are with the Division of Information Science, Graduate School of Science and Technology, Nara Institute of Science and Technology (NAIST), Nara, Japan.}%
\thanks{$^{2}$Y. Saito and K. Ohnishi are with the Haptics Research Center, Keio University, Kanagawa, Japan.}%
\thanks{$^{3}$J. Morimoto is with Graduate School of Informatics, Kyoto University, Kyoto, Japan.}%
\thanks{$^{4}$J. Morimoto is also with the Brain Information Communication Research Laboratory Group (BICR), Advanced Telecommunications Research Institute International (ATR), Kyoto, Japan.}%
}



\maketitle

\begin{abstract}
Robotic grinding is widely used for shaping workpieces in manufacturing, but it remains difficult to automate this process efficiently. In particular, efficiently grinding workpieces of different shapes and material hardness is challenging because removal resistance varies with local contact conditions. Moreover, it is difficult to achieve accurate estimation of removal resistance and analytical modeling of shape transition, and learning-based approaches often require large amounts of training data to cover diverse processing conditions. To address these challenges, we decompose robotic grinding into two components: removal-shape planning and contact-force adaptation. Based on this formulation, we propose DecompGrind, a framework that combines Global Cutting-Surface Planning (GCSP) and Local Contact-Force Adaptation (LCFA). GCSP determines removal shapes through geometric analysis of the current and target shapes without learning, while LCFA learns a contact-force adaptation policy using bilateral control-based imitation learning during the grinding of each removal shape. This decomposition restricts learning to local contact-force adaptation, allowing the policy to be learned from a small number of demonstrations, while handling global shape transition geometrically. Experiments using a robotic grinding system and 3D-printed workpieces demonstrate efficient robotic grinding of workpieces having different shapes and material hardness while maintaining safe levels of contact force.
\end{abstract}

\def\abstractname{Note to Practitioners}
\begin{abstract}
Robotic grinding is widely used in manufacturing to achieve the target shape of a workpiece. Reducing processing time while also preventing excessive contact force is difficult because removal resistance varies with processing conditions, including workpiece shape, material hardness, and contact state. Furthermore, conventional solutions often require complex modeling or learning control policies from a large amount of training data, which is costly to obtain because material removal is irreversible. We propose DecompGrind as a practical solution by separating removal-shape planning from contact-force adaptation. Removal-shape planning is performed by geometric analysis of the current and target shapes without learning, while contact-force adaptation is learned using demonstrations during the grinding of each removal shape. This decomposition reduces the learning burden and allows the policy to be trained from a small number of demonstrations. The proposed method facilitates integration into existing robotic grinding systems and permits deployment in environments where automation has been difficult due to limited training data. It is particularly suitable for environments with diverse workpieces having different shapes and material hardness. However, the method's performance depends on the particular grinding conditions, and substantially different resistance conditions may require additional demonstrations.
\end{abstract}

\begin{figure}[tp]
\centering
\includegraphics[clip,clip,width=.95\columnwidth]{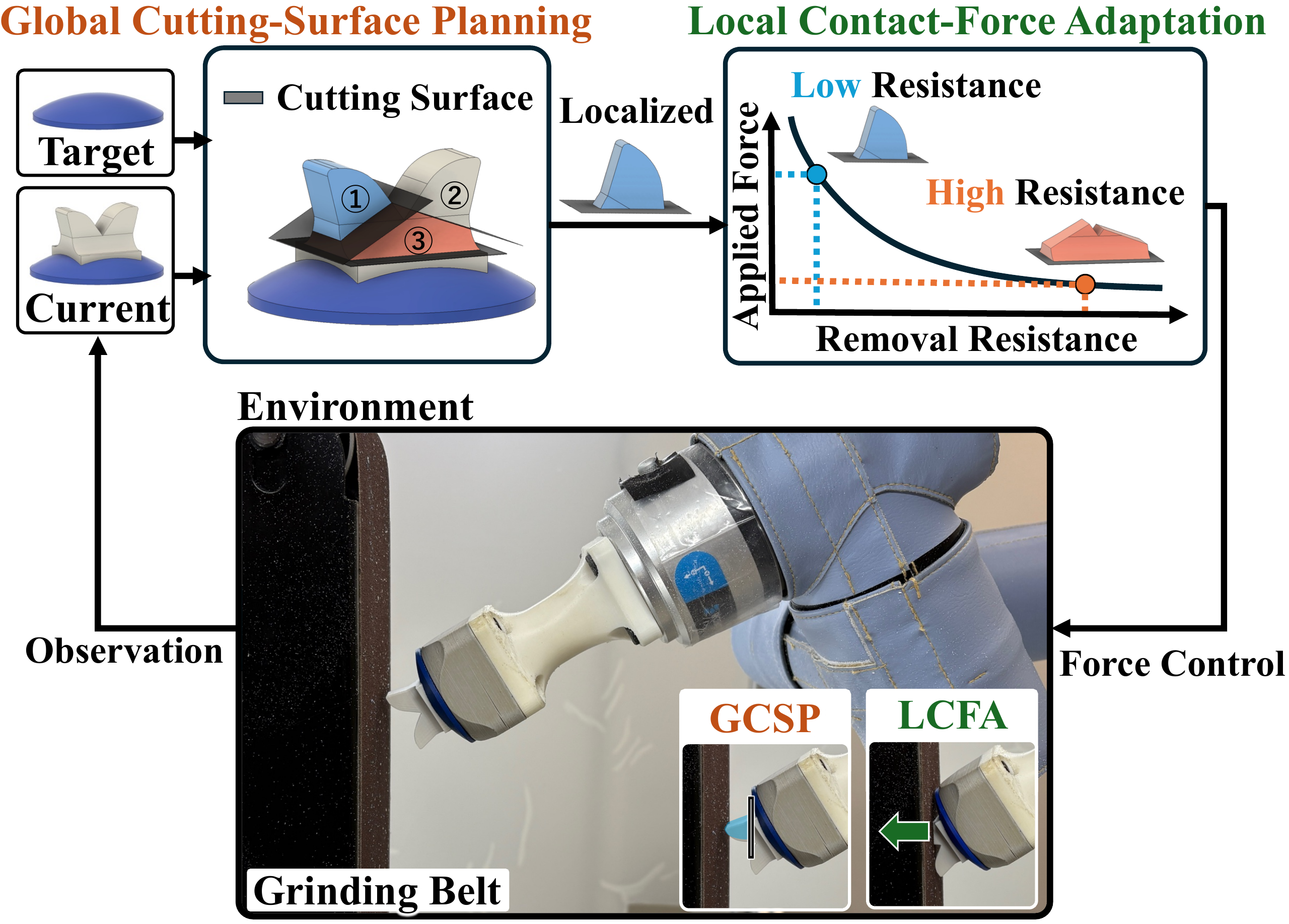}
\caption{Overview of robotic grinding for object shaping with DecompGrind framework.
This framework grinds the workpiece into the target shape while considering removal resistance.
DecompGrind consists of Global Cutting-Surface Planning (GCSP) and Local Contact-Force Adaptation (LCFA).
GCSP is used to plan the next removal shape from the current shape, and LCFA is implemented for grinding each decomposed removal shape with appropriate contact force.
}
\label{fig:eviroment}
\end{figure}

\section{INTRODUCTION}
\IEEEPARstart{M}{achining} is widely used in many industrial fields, such as aerospace and automotive manufacturing \cite{review_grinding_kishore2022comprehensive,review_grinding_zhu2020robotic}.
In machining, grinding is employed for material removal, and its robotic automation has been investigated as a promising extension of the technique \cite{Grinding_Ng,Grinding_Huo}.
Grinding is generally categorized into rough and precision operations depending on the processing objective \cite{review_grinding_WEGENER2017779}.
Rough grinding removes a large amount of material to shape the workpiece, while precision grinding removes a small amount of material to improve surface accuracy.
This paper focuses on robotic rough grinding, in which a workpiece is shaped into the target shape by pressing it against a rotating grinding belt. 
Consequently, in the following, the term grinding refers to rough grinding.

Achieving time-efficient grinding requires the capability to adapt to workpieces having different shapes and material hardness for various tasks.
However, removal resistance in grinding is influenced by complex and only partially observable factors such as local contact conditions, material hardness, and tool-workpiece interaction \cite{stephenson2018metal}.
Accordingly, it clearly becomes difficult to design a model-based force controller and to also determine the appropriate applied force that can cover a variety of different workpieces \cite{TONSHOFF2002551,radu2024review}.
Here, using a large applied force to shorten processing time may damage the robot or the grinding belt.
Conversely, applying constrained grinding to ensure safety increases the processing time. Therefore, a trade-off relationship arises between time-efficiency and safety.

Existing approaches are classified into two categories: geometry-based shape transition modeling \cite{hachimine2023learning,hachimine2024cutting} and data-driven learning from real-robot data \cite{il_force_wang2021robotic,il_force_liu2025forcemimic}. Geometry-based approaches model shape transition effectively but are difficult to apply to force adaptation under varying contact conditions and material hardness.
Data-driven approaches, on the other hand, are promising because they adapt the contact force based on force feedback during grinding. However, they typically require large amounts of training data to learn both shape transition and long-horizon
grinding motions. In grinding tasks, collecting such real-world data is particularly costly because the material removal process is irreversible. Therefore, reducing the learning complexity is crucial for practical robotic grinding systems. 
A framework that facilitates the planning of grinding motion while requiring only limited training data is needed to achieve autonomous object shaping under unpredictable removal resistance.

Grinding involves two fundamentally different processes: global shape evolution and local contact interaction.
Consequently, we formulated grinding as a two-timescale problem to propose a decomposition that separates grinding into (1) global removal-shape planning aimed at a target shape and (2) local contact-force adaptation based on force feedback. The removal-shape planning can be analytically determined from the geometric relationship between the observed and target shapes, while learning is required only for contact-force adaptation during the grinding of each removal shape. This decomposition reduces the learning problem to a local force-adaptation policy, which thus can be trained from a small amount of demonstration data. This formulation converts a long-horizon shaping problem into a short-horizon force adaptation problem that can be learned from a limited number of demonstrations.

Based on the above strategy, this paper proposes DecompGrind, a framework for time-efficient grinding of workpieces having different shapes and material hardness, as illustrated in \figureN{\ref{fig:eviroment}}.
DecompGrind consists of (1) Global Cutting-Surface Planning (GCSP) for determining the removal shape and (2) Local Contact-Force Adaptation (LCFA) based on imitation learning with bilateral control \cite{biadachi2018imitation}.
GCSP models shape transition using a cutting-surface-based geometric representation, allowing the removal shape to be determined geometrically.
LCFA learns the relationship between removal resistance and applied force, permitting force adaptation.
Global shapes encountered in grinding are diverse, and providing shape-specific demonstrations for all cases is impractical; therefore, multiple simple shapes are used to capture variations in removal resistance.
In addition, the learned policy is expected to be applicable to diverse shapes and material hardness, allowing adaptation of the contact force based on removal resistance during grinding of the planned removal shape.
Using a robot, we conducted experiments on 3D-printed workpieces. 
The results show that the proposed method achieves time-efficient grinding with limited training data for imitation learning across different shapes and material hardness.

The main contributions of this paper are as follows:
\begin{itemize}
\item We propose a method that decomposes grinding automation into global removal-shape planning and local contact-force adaptation for time-efficient grinding.
\item We propose DecompGrind, which integrates GCSP for removal-shape generation and LCFA via bilateral control-based imitation learning, thus providing time-efficient grinding with normal force adaptation learned from limited training data.
\item We verified the performance of the proposed method in experiments using real robots, confirming time-efficient grinding of workpieces having different shapes and material hardness.
\end{itemize}

\section{RELATED WORK}
\subsection{Robotic Object-Shape Manipulation}
Many studies have been conducted on object shaping through the removal process based on shape transition  \cite{review_grinding_kishore2022comprehensive,flexible_arriola2020modeling,print_bhatt2020expanding}.
Matl et al.\  \cite{cut_matl2021deformable} and Zhang et al.\  \cite{cut_zhang2025manipulating} learned shape-transition models by reinforcement learning and supervised learning to achieve object shaping toward a target shape.
Hachimine et al.\  \cite{hachimine2023learning,hachimine2024cutting} modeled the removal process as local surface contact and represented shape transition using cutting surfaces for robotic grinding with a grinding belt.
From a different perspective, studies have addressed changes in reaction force during the removal process through force adaptation.
Beltran et al.\  \cite{cut_beltran2024sliceit} learned a velocity-adjustment policy from force feedback through reinforcement learning in a cutting task.
Zhou et al.\  \cite{cut_lyu2025scissorbot} achieved force-adaptive cutting skills using imitation learning from human demonstrators in a paper-cutting task.
Hathaway et al.\ \cite{cut_hathaway2023milling} learned milling strategies under material uncertainty by incorporating a mechanistic cutting-force model.
Shape-transition-based approaches primarily determine geometric motion, such as end-effector position and orientation, but typically they do not explicitly address force adaptation. 
Force-focused studies regulate reaction forces but do not explicitly model shape transition during the removal process.
As a result, existing methods cannot simultaneously model both shape transition and removal resistance, making adaptive robotic grinding under varying removal resistance difficult.

In this study, shape transition in robotic grinding is geometrically represented by cutting surfaces, and removal-shape planning is performed based on this representation.
Furthermore, a policy that imitates a demonstrator’s contact-force adaptation strategy is implemented, and this policy produces grinding that responds to the specific removal resistance of each removal shape.
Consequently, the proposed method allows grinding that accounts for both shape transition and removal resistance.

\subsection{Bilateral Control-Based Imitation Learning for Contact-Force Adaptation}
Demonstrations with real robotic systems are effective for imitation learning in contact-rich tasks, since training data obtained from robotic systems include physical properties such as friction and elasticity that exist in real environments  \cite{il_force_wang2021robotic,il_force_liu2025forcemimic}.
Bilateral control-based imitation learning (BCIL) has been proposed as one approach to imitation learning in contact-rich tasks \cite{biadachi2018imitation,bihieroteical_2022,buamanee2024bi,10883984}.
In this technique, the activity of a demonstrator produces paired position-force data, which can be used for learning the relationship between removal resistance and applied force.
Adachi et al.\  \cite{biadachi2018imitation} used BCIL to maintain a constant contact force in a line-drawing task along a ruler. 
Hayashi et al.\  \cite{bihieroteical_2022} introduced a hierarchical model into BCIL to perform multi-character writing after a small number of demonstrations. 
Kobayashi et al.\  \cite{buamanee2024bi,10883984} incorporated temporal modeling into BCIL using a transformer and visual information for grasping tasks under varying visual conditions.
Conventional BCIL typically learns the target task from demonstrations that cover the entire sequence of actions.
Therefore, in grinding tasks involving diverse conditions, such as shape and material, a large amount of training data is needed, resulting in a high demonstration cost.

In the proposed method, contact-force adaptation is applied only during the grinding of each planned removal shape. 
Although grinding the entire shape as a single continuous process requires a long action sequence, the grinding of each planned removal shape involves a much shorter one. 
Accordingly, only a relatively small amount of training data is required.

\begin{figure}[tp]
\centering
\includegraphics[clip,width=0.99\linewidth]{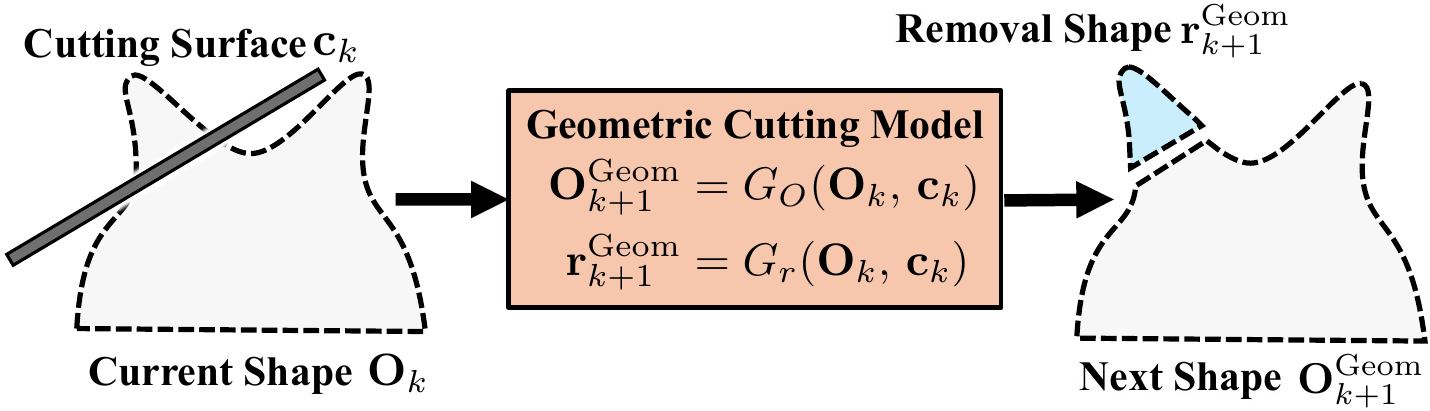}
\caption{Cutting-surface model. The current shape $\mathbf{O}_k$ is divided into a next shape $\mathbf{O}_{k+1}^{\mathrm{Geom}}$ and a removal shape $\mathbf{r}_{k+1}^{\mathrm{Geom}}$ using the cutting surface as a geometric boundary. Here, the superscript $\mathrm{Geom}$ indicates a geometry-based representation.
}
\label{fig:csp}
\end{figure}

\section{Preliminary}
In preparation for explaining the proposed method, Section~\ref{subsec:pre_planner} describes a cutting-surface model used to represent shape transition.
Section~\ref{subsec:pre_contorller} introduces bilateral control-based imitation learning of contact-force adaptation.
These methods serve as the foundation of the proposed GCSP and LCFA.

\subsection{Cutting-Surface Model for Shape Transition}
\label{subsec:pre_planner}
A shape-transition model determines the next shape from the current shape and a particular robot action.
The grinding process is performed through local surface contact between the workpiece and the tool \cite{hachimine2023learning,hachimine2024cutting}.
Without giving consideration to removal resistance, the shape transition is represented geometrically using a cutting surface $\mathbf{c}_{k}$; this representation is referred to as the Geometric Cutting Model (GCM) (\figureN{\ref{fig:csp}}).

The index $k$ denotes the update step of cutting-surface planning.
The cutting surface $\mathbf{c}_{k}$ geometrically divides the current shape $\mathbf{O}_{k}$ into the next (i.e., remaining) shape $\mathbf{O}_{k+1}^{\mathrm{Geom}}$ and the removal shape $\mathbf{r}_{k+1}^{\mathrm{Geom}}$.
This division is defined using the GCM with $G_{O}(\cdot)$ and $G_{r}(\cdot)$ as follows:
\begin{align}
\mathbf{O}_{k+1}^{\mathrm{Geom}} = G_{O}(\mathbf{O}_{k}, \mathbf{c}_{k}),\quad
\mathbf{r}_{k+1}^{\mathrm{Geom}} = G_{r}(\mathbf{O}_{k}, \mathbf{c}_{k}).
\label{eq:geome}
\end{align}
These functions divide the shape geometrically using collision detection without the need for shape-dependent learning.

In actual grinding, removal resistance is generated mainly as the normal force $F^\mathrm{N}_t$ and the tangential force $F^\mathrm{T}_t$ with respect to the grinding belt.
The removal resistance is assumed to depend on the removed volume $V_t$, the belt speed $S_g$, and a coefficient $k_r$ determined from the material type and the belt’s grinding performance.
Accordingly, normal force $F^\mathrm{N}_t$ and tangential force $F^\mathrm{T}_t$ can be approximated as $F^\mathrm{N}_t = k_r V_t / S_g$ and $F^\mathrm{T}_t = \lambda k_r V_t / S_g$, respectively, where $\lambda$ denotes the ratio between the normal and tangential force components \cite{grding_reserch_tang2009modeling}.
Here, it must be noted that the GCM does not consider removal resistance and thus, on its own, cannot determine appropriate actions to handle differences in removal volume and material hardness. This inability can lead to excessive load or increased grinding time.
Consequently, grinding based solely on the GCM reduces operational efficiency.

\begin{figure}[tp]
\centering
\includegraphics[clip,width=0.99\linewidth]{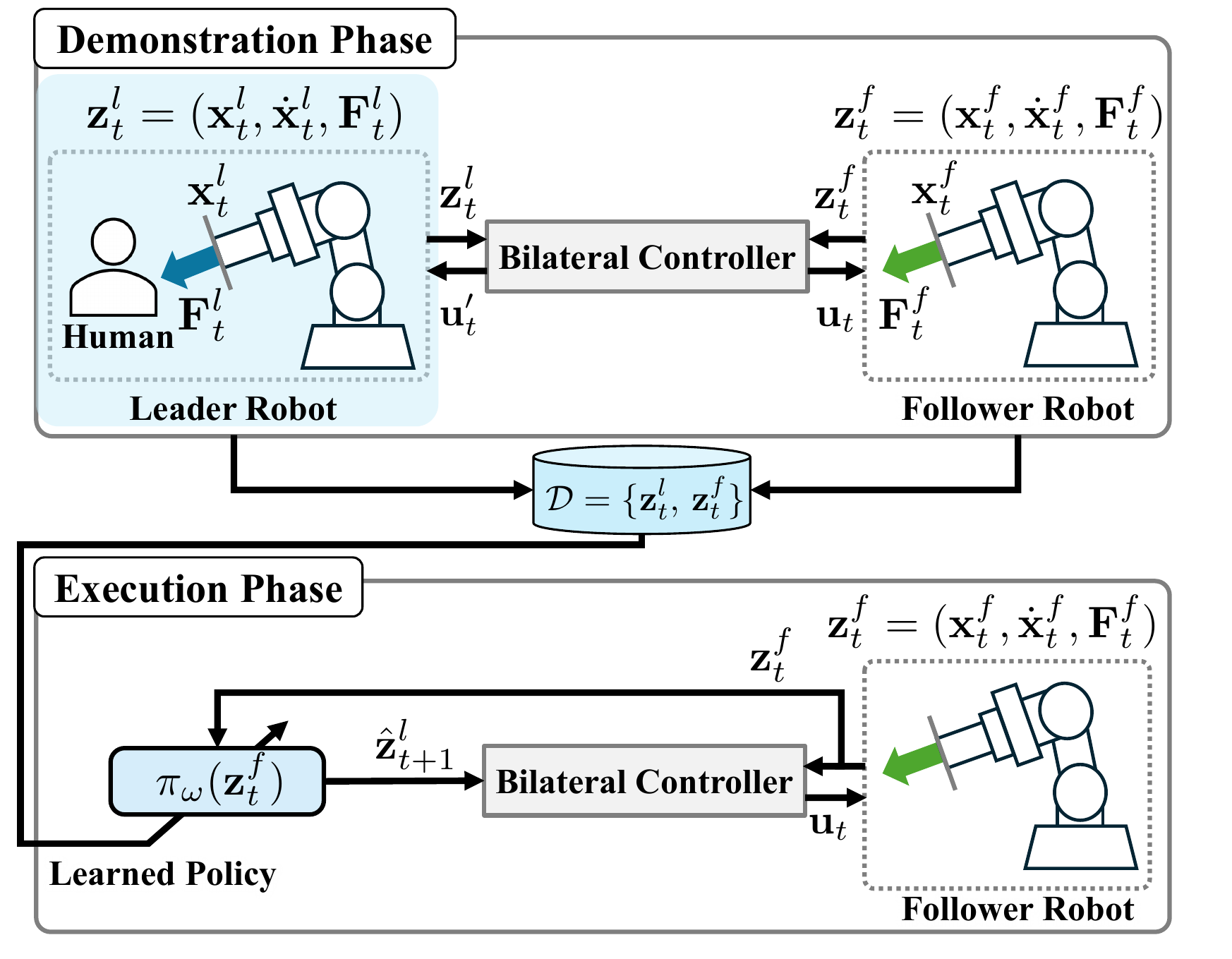}
\caption{Framework of Bilateral Control-based Imitation Learning (BCIL).
Demonstration:
A demonstrator operates the leader robot while the follower robot executes the task via bilateral control, and the robot states are recorded as training data.
Training:
Using the collected training data, the policy is trained to predict the next leader state from the current follower state.
The predicted next leader state is denoted by $\hat{\mathbf{z}}_{t+1}^{l}$.
Execution:
The learned policy substitutes for the leader in bilateral control. The follower robot performs the task accordingly. The policy reproduces the demonstrator’s contact behavior to regulate contact force.
}
\label{fig:bi_framework}
\end{figure}

\subsection{Bilateral Control-Based Imitation Learning of Contact-Force Adaptation }
\label{subsec:pre_contorller}
\subsubsection{Bilateral Control}
In Bilateral Control-based Imitation Learning (BCIL) \cite{biadachi2018imitation}, training data are collected through the use of bilateral control.
Bilateral control applies teleoperation between a leader and a follower robot to achieve position synchronization and force balance.
The control objectives of bilateral control are defined as \cite{5944998}
\begin{align}
\mathbf{x}^l_t - \mathbf{x}^f_t = \mathbf{0}, \quad
\mathbf{F}^l_t + \mathbf{F}^f_t = \mathbf{0}.
\label{eq:bi}
\end{align}

where $\mathbf{x}_t$ and $\mathbf{F}_t$ denote the position and force vectors of the robot end-effector, and $l$ and $f$ represent the leader and follower robot systems.
Here, $t$ denotes the time step.
Equation \eqref{eq:bi} enforces the position synchronization and action-reaction force relation between the systems, enabling the demonstrator to manipulate the robot while perceiving haptic feedback.

In studies using bilateral control  \cite{biadachi2018imitation,5944998}, a hybrid controller that combines position and force control is commonly used to control the robot.
\begin{align}
\label{eq:hybrid}
&\mathbf{u}_t = \nonumber \\
&\frac{1}{2}\mathbf{J}\left[\mathbf{K}_p(\mathbf{x}_t^l-\mathbf{x}_t^f)
+\mathbf{K}_d(\dot{\mathbf{x}}_t^l-\dot{\mathbf{x}}_t^f)\right]
+\frac{1}{2}\mathbf{K}_f(\mathbf{F}_t^l+\mathbf{F}_t^f).
\end{align}

Here, $\mathbf{u}_t$ denotes the robot control input in the force dimension, $\dot{\mathbf{x}}_t$ the end-effector velocity, $\mathbf{J}$ the inertia matrix, and $\mathbf{K}_p$, $\mathbf{K}_d$, and $\mathbf{K}_f$ denote diagonal gain matrices for position, velocity, and force control, respectively.
This controller scheme enables the demonstrator to provide demonstrations using both position and force while receiving haptic feedback.

\subsubsection{Bilateral Control-Based Imitation Learning for Robotic Grinding}
The framework of BCIL is shown in \figureN{\ref{fig:bi_framework}}.
The state $\mathbf{z}_t$ required for bilateral control consists of the robot position $\mathbf{x}_t$, velocity $\dot{\mathbf{x}}_t$, and force $\mathbf{F}_t$ vectors.
In the demonstration phase, the demonstrator manipulates the leader robot while the follower robot executes the task, from which training data are collected.
Using the collected data, the parameters $w$ of a policy $\pi_w$ are learned to estimate the next leader-robot contact state $\hat{\mathbf{z}}_{t+1}^{l}$ from the follower-robot contact state $\mathbf{z}_{t}^{f}$.
In the execution phase, the leader-robot operation is replaced by the learned policy $\pi_w$  \cite{biadachi2018imitation}.
Consequently, the learned policy substitutes for the leader and permits the execution of contact-rich tasks.

\section{PROPOSED METHOD}
This section describes the formulation of the grinding task and the proposed DecompGrind method.
Section~\ref{subsec:pre_model} describes the state transition in grinding and decomposes grinding actions into global removal-shape planning and local contact-force adaptation.
Section~\ref{subsec:proposed_overview} introduces DecompGrind.
The remainder of this section describes in greater detail the two components of DecompGrind: Global Cutting-Surface Planning (GCSP, Section~\ref{subsec:cuttingsurface}), 
and Local Contact-Force Adaptation (LCFA, Section~\ref{subsec:grindingadustment}).
In this framework, the planning module GCSP operates at the planning step $k$, 
while the execution module LCFA operates at the time step $t$.

\subsection{State Transition in Grinding}
\label{subsec:pre_model}
\subsubsection{Problem Setting}
Grinding is the process of removing material from an initial shape toward forming the target shape.
The variable $t$ denotes the time step, and the state in the grinding task $\mathbf{s}_t$ consists of the current shape $\mathbf{O}_t$ and the robot state $\mathbf{x}_t$:
\begin{align}
  \mathbf{s}_t = (\mathbf{O}_t,\mathbf{x}_t).
\end{align}
The next state $\mathbf{s}_{t+1}$ is determined by the current state $\mathbf{s}_t$ and the action $\mathbf{a}_t$, and the state-transition model $f$ is defined as follows:
\begin{align}
  \mathbf{s}_{t+1} = f(\mathbf{s}_t,\mathbf{a}_t).
\end{align}
Here, the action $\mathbf{a}_t$ represents the grinding motion executed by the robot (e.g., end-effector pose and contact force).
The policy that determines the action from the state is defined as follows:
\begin{align}
  \mathbf{a}_t = \pi(\mathbf{s}_t).
\end{align}

In grinding tasks, modeling state transitions under the condition of removal resistance is difficult, and obtaining the data needed to learn removal resistance for diverse shapes and material hardness is also challenging.
Therefore, both analyzing the transition $f$ and learning the policy $\pi$ are difficult.

\subsubsection{Decomposition of Grinding Process Based on Contact State}
The removal process in grinding is performed under local contact conditions between the workpiece and the tool surface.
Grinding can be decomposed into planning the removal shape $\mathbf{r}_{t+1}^\mathrm{Geom}$ and adapting the pressing force $\mathbf{u}_t$ to execute the removal.
The robot action $\mathbf{a}_t$ is decomposed into two components: the cutting surface $\mathbf{c}_t$, which determines the removal shape $\mathbf{r}_{t+1}^\mathrm{Geom}$, and the pressing force $\mathbf{u}_t$ as follows:
\begin{align}
\label{eq:dividepolicy}
\mathbf{a}_t = (\mathbf{c}_t,\mathbf{u}_t)
=(\pi^{\mathrm {place}}(\mathbf{O}_t),\pi^{\mathrm {force}}_w(\mathbf{z}_t)).
\end{align}
Here, $\pi^{\mathrm{place}}$ is the policy that determines the cutting surface $\mathbf{c}_t$ from the shape $\mathbf{O}_t$, and this can be planned geometrically by the cutting-surface model without learning. In addition, $\pi^{\mathrm{force}}_w$ is the policy that determines the pressing force $\mathbf{u}_t$ from the contact state $\mathbf{z}_t$. This decomposition reduces the learning target to the shape-independent force-adaptation policy $\pi^{\mathrm{force}}_w$, thus simplifying the learning problem.

\subsubsection{Decomposition into Shape Transition and Force Adaptation}
In this work, the grinding task is decomposed into 
global removal-shape planning and local contact-force adaptation.
Because these components play different roles and are updated at different frequencies, two time indices are introduced: 
a planning step $k$ and a time step $t$.

In global removal-shape planning, a cutting surface $\mathbf{c}_k$ 
is determined at each planning step $k$, and the workpiece shape 
is updated geometrically; therefore, the shape transition is formulated 
independently of the robot state.
Using the GCM \equationname{\ref{eq:geome}}, the shape transition is given by
\begin{align}
\mathbf{O}_{k+1} := \mathbf{O}_{k+1}^\mathrm{Geom} = G_O(\mathbf{O}_k, \mathbf{c}_k).
\end{align}

In local contact-force adaptation, the cutting surface $\mathbf{c}_k$ 
is treated as fixed within planning step $k$, and the robot state 
is updated at time step $t$ according to the applied force $\mathbf{u}_{t}$:
\begin{align}
\mathbf{x}_{t+1}
= f_x(\mathbf{x}_{t}, \mathbf{u}_{t}, \mathbf{c}_k).
\end{align}
Here, $f_x$ denotes the robot-state transition function given the applied force.
Consequently, the shape transition can be described purely by geometric quantities,
and the robot-state transition can be treated within a common framework 
regardless of shape differences.
This formulation is motivated by the observation that removal resistance in grinding is primarily determined by local contact interactions rather than the global workpiece shape. Without such decomposition, the policy would need
to learn both shape transition and force regulation jointly. In contrast, the proposed decomposition isolates the learning component to local force adaptation, allowing the policy to be trained by a limited number of demonstrations.

\begin{figure*}[t]
\centering
\includegraphics[clip,width=0.89\linewidth]{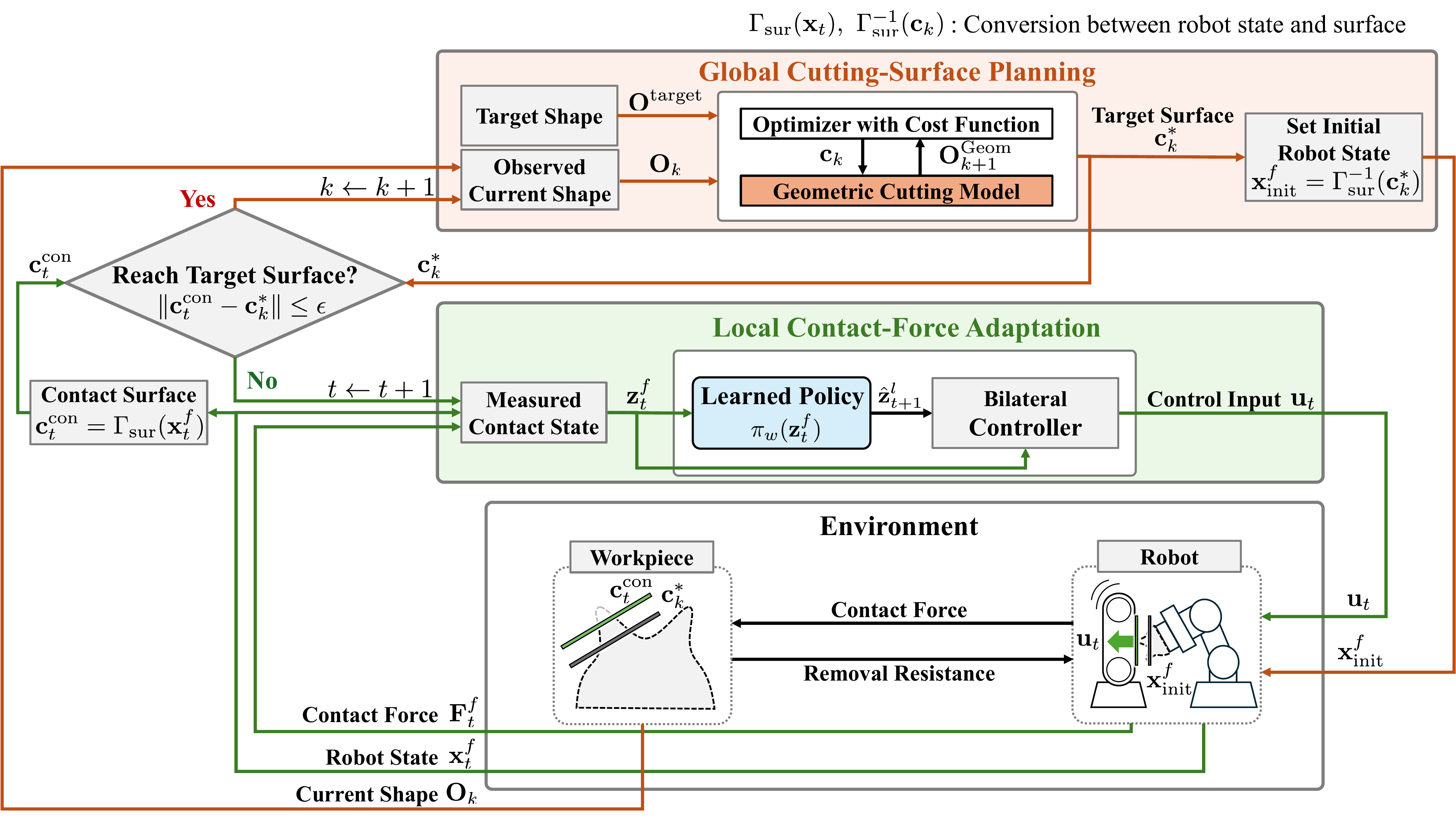}
\caption{Framework of DecompGrind.
A cutting surface is planned from the current and target shapes.
The robot is then set up at the initial position based on the planned cutting surface.
Next, the learned LCFA is then used to grind the removal shape defined by GCSP.
Grinding continues until the contact surface $\mathbf{c}_t^{\mathrm{con}}$ reaches the cutting surface $\mathbf{c}_k^*$.
Here, $\mathbf{c}_t^{\mathrm{con}}$ denotes the contact surface between the belt and the workpiece, 
$\mathbf{c}_k$ denotes a candidate cutting surface, 
and $\mathbf{c}_k^*$ denotes the selected optimal cutting surface.
When $\|\mathbf{c}_t^\mathrm{con}-\mathbf{c}_k^*\|<\epsilon$, the current shape is observed and the next cutting surface is planned.
Here, $\epsilon>0$ is a small positive constant.
The index $k$ denotes the update step of GCSP and $t$ denotes the time step of LCFA.
The functions $\Gamma_{\mathrm{sur}}(\cdot)$ and $\Gamma_{\mathrm{sur}}^{-1}(\cdot)$ represent the conversion between the robot state and the surface representation.}
\label{fig:Framework}
\end{figure*}

\subsection{DecompGrind}

\label{subsec:proposed_overview}

Based on the formulation in Section~\ref{subsec:pre_model}, we derive DecompGrind (\figureN{\ref{fig:Framework}}).
DecompGrind consists of a Global Cutting-Surface Planning (GCSP) and a Local Contact-Force Adaptation (LCFA).
GCSP generates a cutting surface based on the current shape,
and LCFA is used to grind the planned surface by regulating the applied force.
This process is repeated until the target shape error falls below a predefined threshold.

\subsubsection{Global Cutting-Surface Planning (GCSP)}
GCSP selects an optimal cutting surface using the cutting-surface model to globally determine the removal shape toward forming the target shape.
Details are provided in Section~\ref{subsec:cuttingsurface}.

\subsubsection{Local Contact-Force Adaptation (LCFA)}
While GCSP determines the geometric removal sequence, it cannot regulate the contact force during grinding.
Without appropriate force adaptation, excessive force can occur when the contact area becomes large or the material hardness is high, leading to unstable grinding or longer grinding time.
LCFA adapts the contact force for the geometrically decomposed local shapes determined by GCSP, and it can be learned from a small amount of data through demonstrations of force adaptation during grinding.
Details are provided in Section~\ref{subsec:grindingadustment}.

By combining these two components, grinding automation is achieved.
First, demonstrations are conducted on simple shapes in advance, and the relationship between removal resistance and contact force is learned using bilateral control.
Next, based on the observed shape, the removal shape toward forming the target shape is determined by GCSP.
The removal shape is then processed using the learned force policy until the cutting surface is reached.
After such grinding, the shape is observed again, and this process is repeated until the target shape error falls below a threshold.

\subsection{Global Cutting-Surface Planning (GCSP)}
\label{subsec:cuttingsurface}

GCSP determines the optimal cutting surface $\mathbf{c}_k^*$ from the observed shape $\mathbf{O}_k$ and the target shape $\mathbf{O}^{\mathrm{target}}$.
The shape after grinding influences subsequent cutting-surface selection, requiring GCSP to account for future shape evolution.
Using the geometric cutting model $G_O$ in \equationname{\ref{eq:geome}}, the problem is formulated as an optimization that obtains a sequence of cutting surfaces within a model predictive control (MPC) framework \cite{garcia1989model}:
\begin{equation}
  \label{eq:csp}
  \left.
  \begin{alignedat}{2}
    \mathbf{c}_{k:k+H-1}^*
     = \operatorname*{arg\,min}_{\mathbf{c}_{k:k+H-1}}
        \frac{1}{H}\sum_{h=k}^{k+H-1} C(\mathbf{O}^{\mathrm{target}},\mathbf{O}_h,\mathbf{c}_h),\\
    \text{subject to}\quad
    \mathbf{O}_{h+1} = G_O(\mathbf{O}_h,\mathbf{c}_h).
  \end{alignedat}
  \right\}
\end{equation}

Here, $H$ denotes the planning horizon and $h$ denotes the planning index.
The cost function $C(\mathbf{O}^{\mathrm{target}},\mathbf{O}_h,\mathbf{c}_h)$ is defined as the sum of the target-shape error and the removal-shape cost, and thus the global shape error and local removal characteristics are considered simultaneously.
After determining the cutting surface, the initial robot state 
$\mathbf{x}_{\mathrm{init}}^f$ is computed from the planned cutting surface as 
$\mathbf{x}_{\mathrm{init}}^f=\Gamma_{\mathrm{sur}}^{-1}(\mathbf{c}_h)$,
where $\Gamma_{\mathrm{sur}}^{-1}(\cdot)$ computes the robot state corresponding to the cutting surface.
The robot is then moved to this initial state before executing the grinding motion.

\subsection{Local Contact-Force Adaptation (LCFA)}
\label{subsec:grindingadustment}
In belt grinding, removal resistance mainly appears as tangential and normal forces,
where the tangential force depends on the applied normal force \cite{grding_reserch_tang2009modeling}.
Accordingly, for each local shape determined by GCSP, LCFA learns to adjust the applied normal force based on the tangential force.
Grinding of the removal shape $\mathbf{r}_{k+1}^\mathrm{Geom}$ is completed when the contact surface between the belt and the workpiece reaches the cutting surface $\mathbf{c}_k^*$.
Thus, grinding continues until the following condition is satisfied:
\begin{equation}
\|\mathbf{c}_t^{\mathrm{con}}-\mathbf{c}_k^*\| < \epsilon .
\label{eq:bi_error}
\end{equation}
Here, $\epsilon>0$ is a small positive constant, and the contact surface 
$\mathbf{c}_t^{\mathrm{con}}$ is geometrically computed from the follower robot's state as 
$\mathbf{c}_t^{\mathrm{con}}=\Gamma_{\mathrm{sur}}(\mathbf{x}_t^f)$,
where $\Gamma_{\mathrm{sur}}(\cdot)$ computes the contact surface corresponding to the robot state.

LCFA is learned using BCIL to model the relationship between the tangential force and the applied normal force.
In the demonstration phase, the applied normal force is not constant but instead adaptively modulated according to the tangential force.
The policy is defined to estimate the next leader robot's contact state $\mathbf{z}_{t+1}^{l}$ from the sequence of follower robot's contact states over the most recent $n\in\mathbb{N}$ steps, $\mathbf{z}_{t-n+1:t}^{f}$, thus capturing dynamic force responses.
Using the training dataset $\mathcal{D}={(\mathbf{z}^{f}_{t-n+1:t},\mathbf{z}^{l}_{t+1})}_{t=n}^{T-1}$, we adopt the following least-squares formulation to learn the policy $\pi_{w}$:
\begin{equation}
\begin{split}
w^{*} =
\arg\min_{w}
\mathbb{E}_{(\mathbf{z}^f,\mathbf{z}^l)\sim\mathcal{D}}
\Big[
\sum_{t=n}^{T-1}
\left\|
\mathbf{z}_{t+1}^{l}
-
\pi_{w}(\mathbf{z}_{t-n+1:t}^{f})
\right\|_2^2
\Big].
\end{split}
\label{eq:window_imitation}
\end{equation}

Here, $T$ denotes the length of each demonstration episode.
In the execution phase, the control input $\mathbf{u}_t$ of the follower robot is computed by substituting the leader robot's contact state $\mathbf{z}_t^l$ in \equationname{\ref{eq:hybrid}} with the estimated state $\hat{\mathbf{z}}_{t+1}^l$ predicted by the policy $\pi_{w^{*}}$.

\begin{figure}[tp]
\centering
\includegraphics[clip,width=0.9\linewidth]{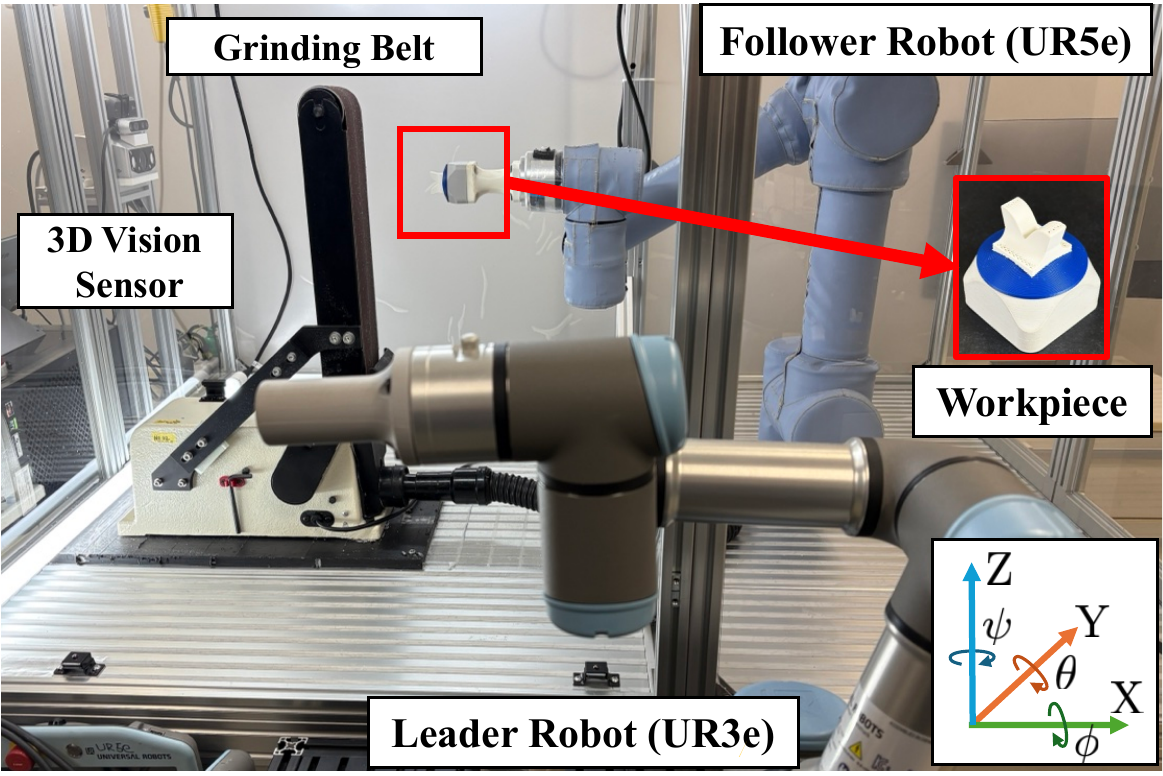}
\caption{Configuration of grinding environment.
The experimental setup consists of a leader-follower robot system
connected via bilateral control.
The workpiece is attached to the follower end-effector
and observed with a 3D vision sensor.
During the demonstration phase,
the demonstrator manipulates the leader robot
under contact-force feedback from the environment,
and the follower robot performs grinding through bilateral control.
In the execution phase,
the follower robot performs grinding
based on the policy learned via BCIL,
without using the leader robot.
}
\label{fig:env}
\end{figure}

\section{Rough Grinding Environment for Robots}

As shown in \figureN{\ref{fig:env}}, two 6-DOF manipulators (UR3e and UR5e) are used as the leader and follower robots.
The grinding belt is fixed in the environment, and the workpiece is attached to the follower end-effector.
Shape observation is performed using a 3D vision sensor (YCAM3D-10M). The observed point-cloud data are used to represent the current shape in the proposed framework.

The workpieces used in the experiments are fabricated with a 3D printer and, as shown in \tableN{\ref{tab:shape}}, consist of three parts: the base (gray), the target shape (blue), and the material removed by grinding (white).
These workpieces differ in shape and density, where the density variation is used to emulate differences in material hardness, allowing the system to be evaluated under diverse processing conditions.

\section{EXPERIMENT}

To evaluate whether the proposed decomposition framework provides efficient grinding while also limiting the required amount of training data, the following research questions are addressed:
\begin{description}
\item[RQ1:] How does the proposed decomposition framework achieve efficient grinding of workpieces with different shapes and material hardness using limited training data? (\ref{subsec:resultpropose})
\item[RQ2:] How does LCFA  adapt the contact force to changes in removal resistance caused by differences in shape and material hardness? (\ref{subsec:simpleshape})
\item[RQ3:] How does LCFA improve the time required to reach a predefined target shape error threshold? (\ref{subsec:resultfai})
\item[RQ4:] How does the selection of workpieces for training data collection affect the learning performance of LCFA policy? (\ref{subsec:resultteachingshape})
\end{description}

\begin{table}[tp]
\centering
\caption{WorkPiece for Evaluation (WP-E): \textnormal{Initial and target shapes}}
\begin{tabular}{cccc}
\toprule
Name & WP-E1 & WP-E2 & WP-E3\\
 \midrule
\raisebox{4\height}{Initial}  &
\includegraphics[width=1.69cm]{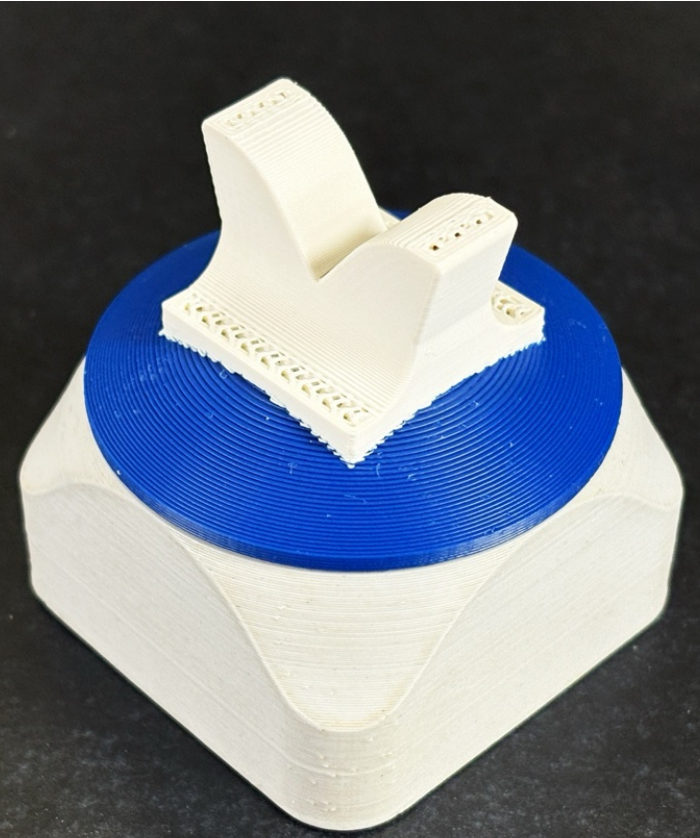} & 
\includegraphics[width=1.8cm]{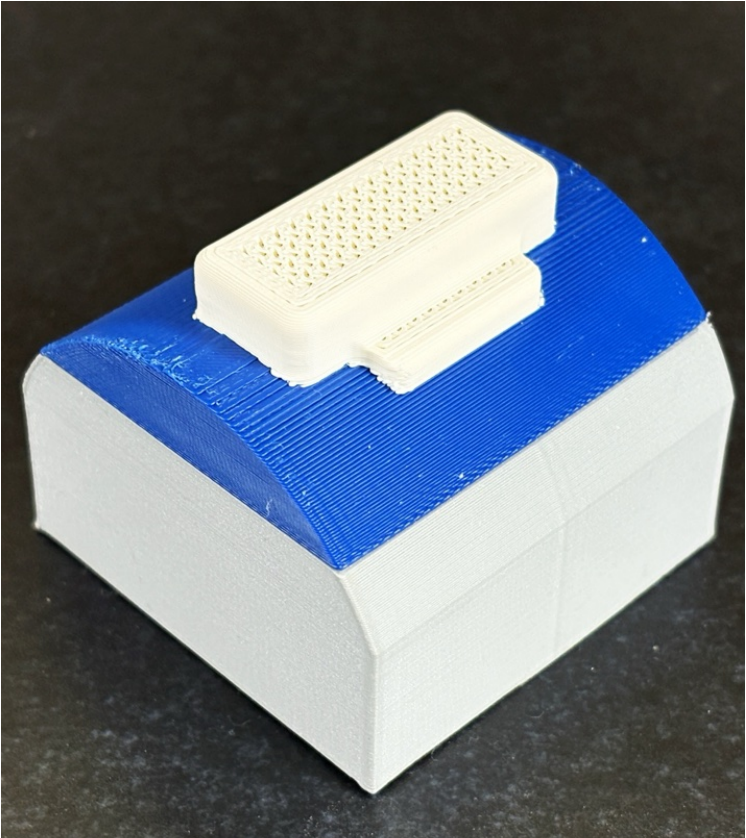} &
\includegraphics[width=1.675cm]{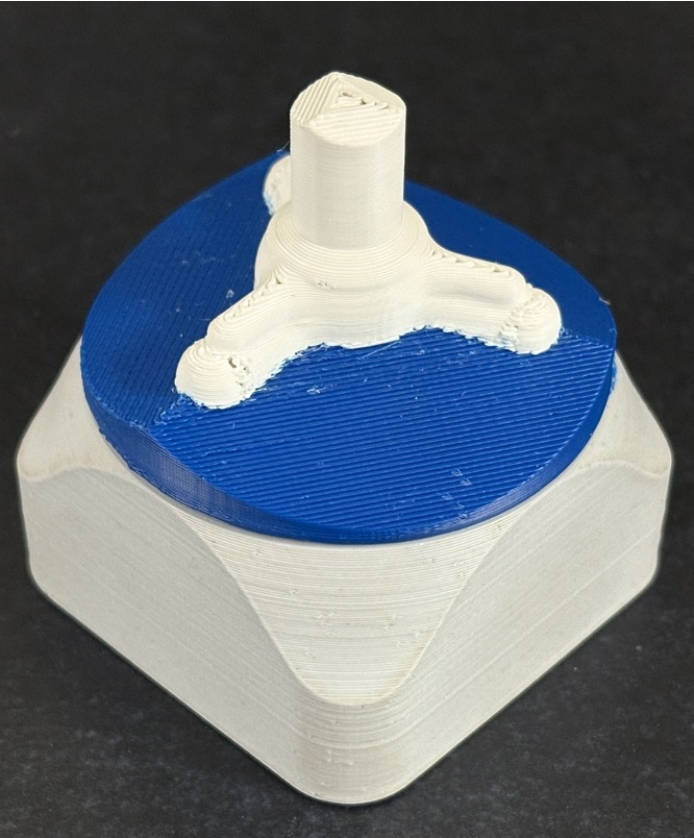} \\
 \midrule
\raisebox{4\height}{Target}  &
\includegraphics[width=1.69cm]{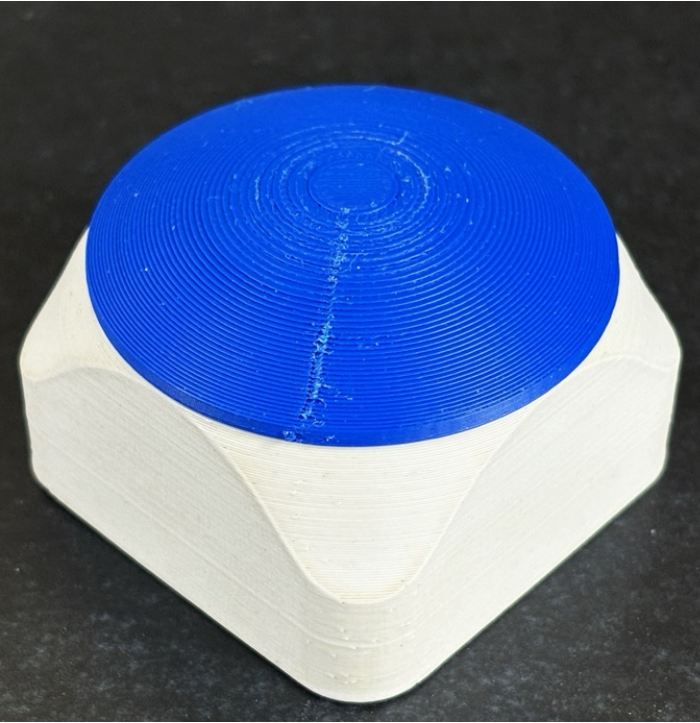} &
\includegraphics[width=1.8cm]{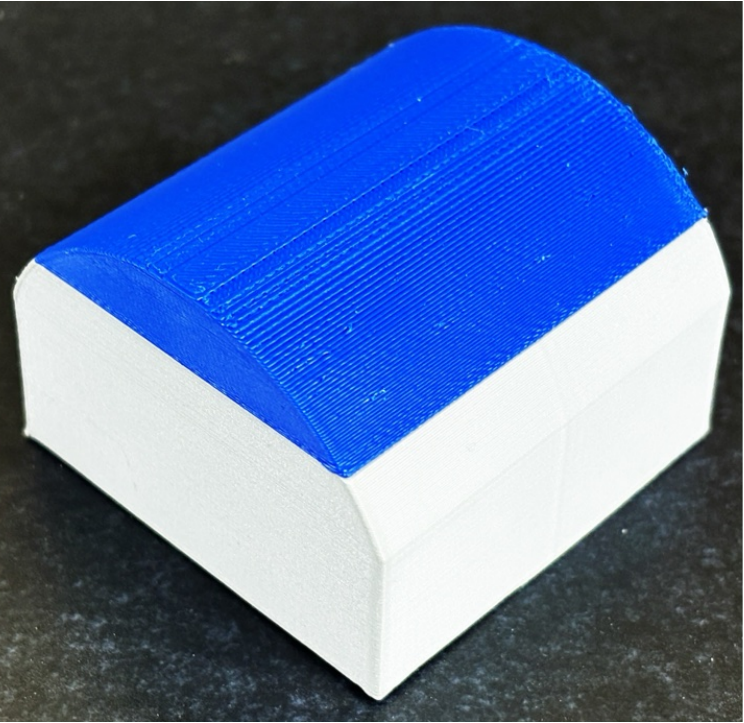} &
\includegraphics[width=1.69cm]{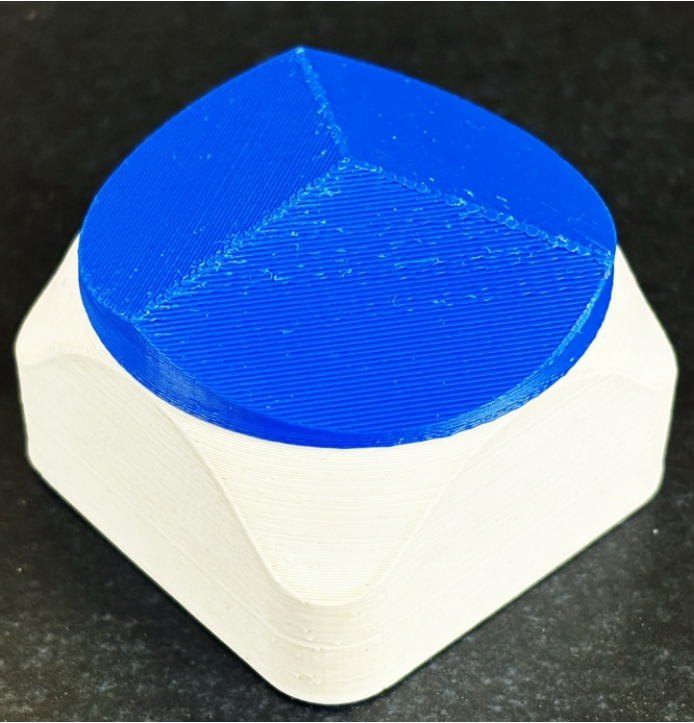} \\
 \midrule
Density [$\%$] & 30 & 45 & 60 \\
\bottomrule
\end{tabular}
\label{tab:shape}
\end{table}

\subsection{Experimental Setup}

A force limit was enforced to prevent damage to the robot and the grinding belt: Whenever the tangential force ($\mathrm{Z}$-axis) exceeded 9 N, the current grinding process was terminated. This threshold was determined empirically to ensure safe operation of the robot and the grinding belt. For the cutting-surface-based methods, the process was restarted from the shape observation step. Grinding experiments were repeated three times for each method and each workpiece.

\subsection{Settings for GCSP}

\subsubsection{Observation and Shape Representation}
The observed shape is represented as a point cloud $\mathbf{O}_k = [q_{i,k}]_{i=1}^{D}$,
where $D$ denotes the number of points and $q_{i,k}$ is the 3D position of particle $i$ at step $k$.
Geometric processing and visualization of the point cloud are performed with Open3D  \cite{zhou2018open3d} and PyVista  \cite{sullivan2019pyvista}.

\subsubsection{Representation and Update of Cutting Surface}
Under the assumption that the grinding belt surface is sufficiently wide, the cutting surface $\mathbf{c}_k$ is independent of the two translational DOFs and one rotational DOF around the vertical axis, and it is represented as $\mathbf{c}_k = [\theta_k,\psi_k, x_k]$.
In the experiments, the workpiece shape was observed every two planning steps (i.e., every two increments of $k$),
and the subsequent removal shape $\mathbf{r}_{k+1}^\mathrm{Geom}$ was replanned accordingly.

\begin{figure}[t]
\centering
\includegraphics[width=0.9\linewidth]{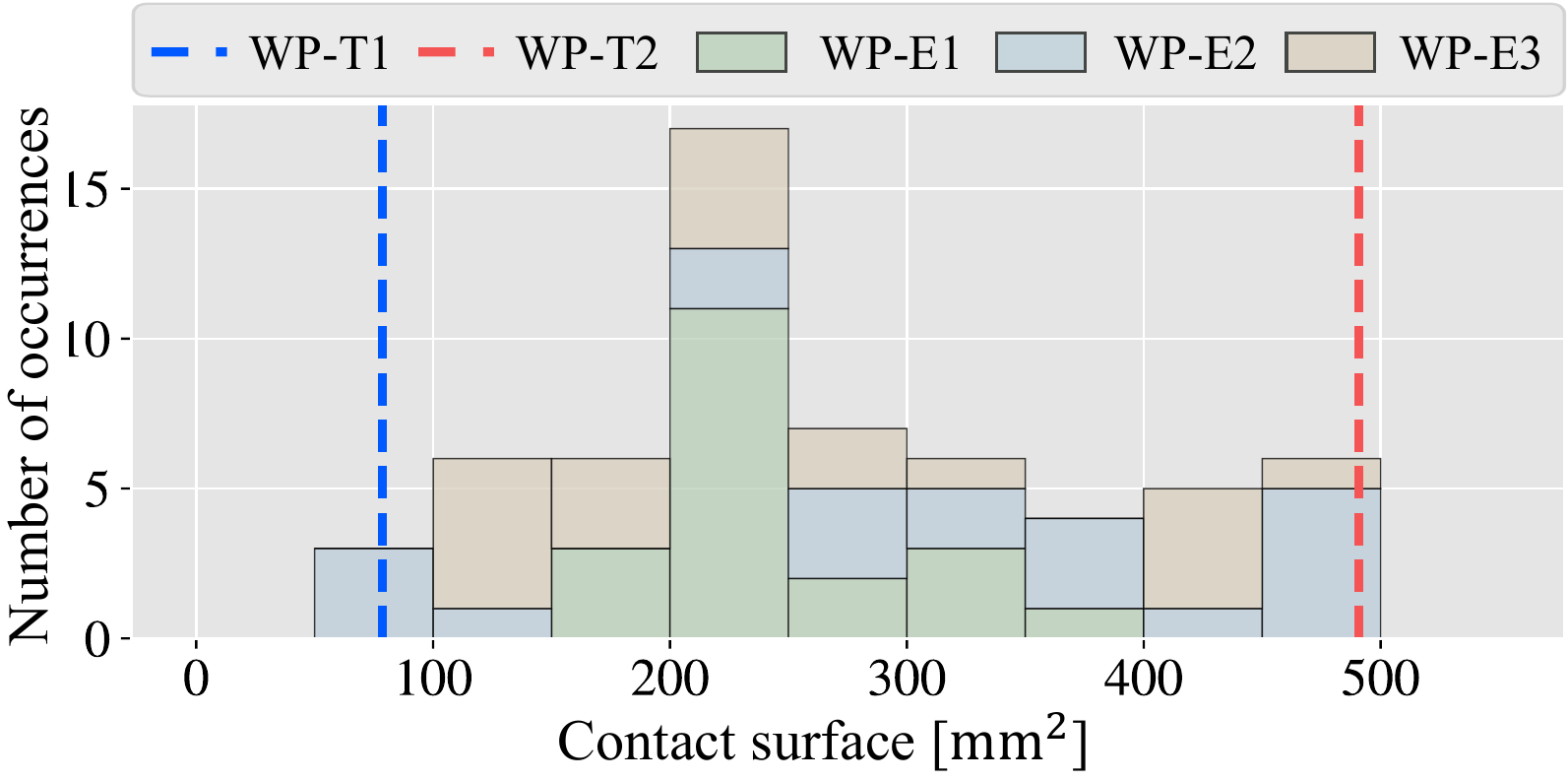}
\caption{Analysis of contact surface distribution with GCSP in a simulated environment}
\label{fig:shape_analyze}

\end{figure}
\begin{table}[t]
\centering
\caption{WorkPiece for LCFA Training (WP-T): \textnormal{Different densities, diameters, and heights. WP-T1 corresponds to blue dashed line in \figureN{\ref{fig:shape_analyze}}, and WP-T2 corresponds to red dashed line.}}
\begin{tabular}{ccc}
\toprule
Name & WP-T1 & WP-T2\\
 \midrule
\raisebox{4\height}{Initial}  &
\includegraphics[width=1.8cm]{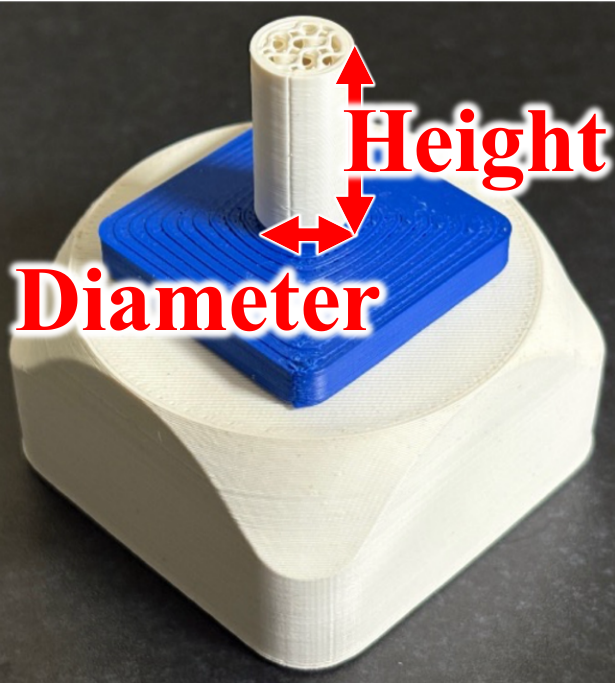} & 
\includegraphics[width=1.8cm]{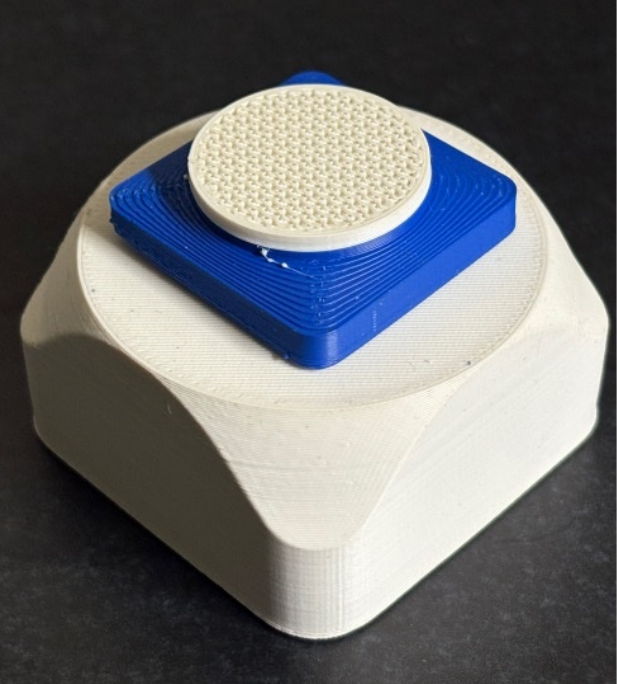} \\
 \midrule
Density [$\%$] & 30& 60\\
Diameter [mm] & 10  & 25  \\
Height [mm] & 20 & 2 \\
\bottomrule
\end{tabular}
\label{tab:teach_shape}
\end{table}

\setlength{\tabcolsep}{1pt}
\begin{table}[tp]
\centering
\caption{WorkPiece with Simple Shape (WP-S): \textnormal{Different density, diameter, and height}}
\label{tab:simpleshape}
\begin{tabular}{ccccccc}
\toprule
Name & WP-S1 & WP-S2 & WP-S3 & WP-S4 & WP-S5 \\
 \midrule
\raisebox{3\height}{Initial}  &
\includegraphics[width=1.35cm]{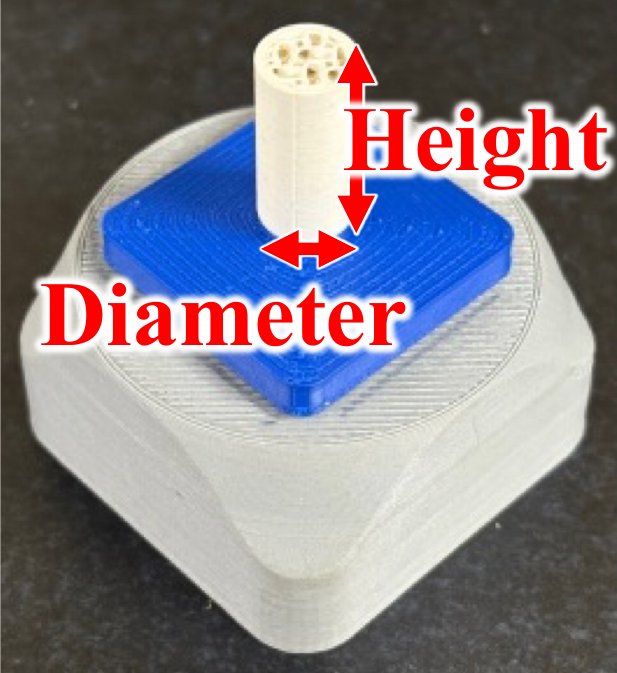} & 
\includegraphics[width=1.34cm]{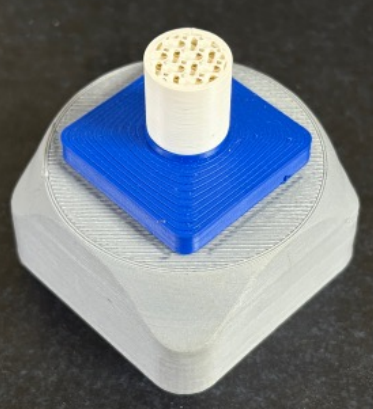} & 
\includegraphics[width=1.33cm]{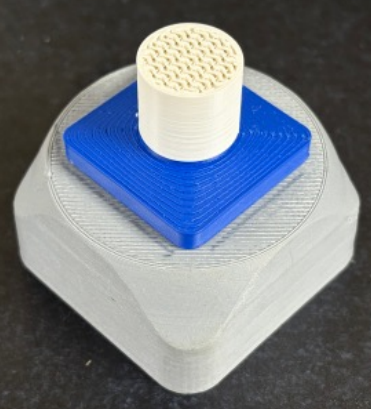} & 
\includegraphics[width=1.34cm]{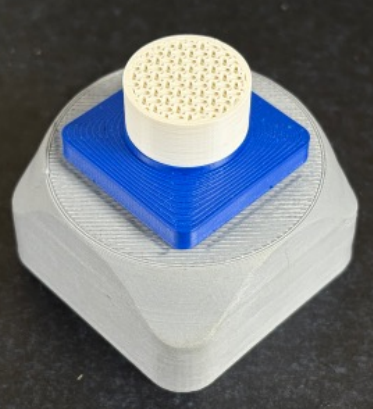} & 
\includegraphics[width=1.34cm]{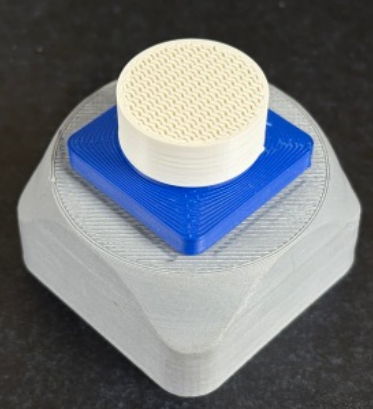} \\
 \midrule
Density [$\%$] & 30 & 30 & 45 & 45 & 60 \\
Diameter [mm] & 10 & 14 & 17 & 21 & 25 \\
Height [mm]& 20 & 15 & 15 & 10 & 10 \\
\bottomrule
\end{tabular}
\end{table}

\subsubsection{Cost Function for GCSP}
The cost function used in \equationname{\ref{eq:csp}} is defined as
\begin{align}
  \label{eq:cost}
    C(\mathbf{O}_h,\mathbf{c}_h)=d_{CD}(\mathbf{O}_h,\mathbf{O}^{\mathrm{target}})+\eta(\mathbf{O}_h,\mathbf{c}_h).
\end{align}
The first term denotes the shape error, and the second term denotes the geometric characteristics of the removal shape.

The first term evaluates the difference between two point-cloud datasets using the Chamfer discrepancy  \cite{nguyen2021point}.
\begin{align}
    &d_{CD}(\mathbf{O}_1,\mathbf{O}_2) = \nonumber\\
    &\frac{1}{|\mathbf{O}_1|}\sum_{p\in \mathbf{O}_1}\underset{q\in \mathbf{O}_2}{\operatorname{min}} \|\mathbf{p}-\mathbf{q}\|^2_2 + \frac{1}{|\mathbf{O}_2|}\sum_{q\in \mathbf{O}_2}\underset{p\in \mathbf{O}_1}{\operatorname{min}} \|\mathbf{p}-\mathbf{q}\|^2_2.
    \label{eq:error}
\end{align}

The second term represents the property by which a narrow and elongated removal shape allows stronger and faster grinding than a wide and shallow shape.
Here, $V(\mathbf{O}_h,\mathbf{c}_h)$ denotes the removed volume, $h(\mathbf{O}_h,\mathbf{c}_h)$ the vertical height, and $k_c$ a proportional constant.
The second term is defined as follows:
\begin{align}
    \eta(\mathbf{O}_h,\mathbf{c}_h) = k_c V(\mathbf{O}_h,\mathbf{c}_h) / h(\mathbf{O}_h,\mathbf{c}_h).
\end{align}

\subsection{Settings for LCFA}

\subsubsection{Control Axes During Demonstration and Learning}

In grinding using cutting surfaces, after aligning the end-effector orientation parallel to the cutting surface, grinding can be performed by moving only along the normal direction ($\mathrm{X}$-axis).
However, removal resistance acts not only in the normal direction
but also strongly in the tangential direction \cite{grding_reserch_tang2009modeling}.
Therefore, the use of only a single axis makes stable demonstrations difficult.
In this experiment, demonstrations and learning were performed in both
the normal and tangential directions.

Grinding of the removal shape $\mathbf{r}_{k+1}^\mathrm{Geom}$ is completed when the contact surface $\mathbf{c}_t^{con}$ reaches the cutting surface $\mathbf{c}_k^*$.
Since grinding proceeds along the normal direction ($\mathrm{X}$-axis), the error in the termination condition (\ref{eq:bi_error}) is evaluated only along this direction.
The parameter $\epsilon$ in (\ref{eq:bi_error}) was set to 0.05\,mm, and grinding was terminated when the condition was satisfied in more than 10 control steps.

Under these settings, the control gains in (\ref{eq:hybrid}) were set as
$K_p = 360$, $K_d = 120$, and $K_f = 1$ to control each axis.

\subsubsection{Selection of Workpieces for Training Data Collection}
\label{subsec:segmented_analyze}
LCFA learns the relationship between the tangential force and the applied normal force for each removal shape.
Training data are collected using bilateral control on simple shapes that facilitate stable force adaptation.
GCSP was applied to each workpiece listed in \tableN{\ref{tab:shape}} to analyze trends in the cutting-surface area for selecting the workpieces to use in collecting training data.
The analysis results for each workpiece are shown in \figureN{\ref{fig:shape_analyze}}.
Using the minimum and maximum base areas obtained from this analysis, simple workpieces with matched base areas were designed as shown in \tableN{\ref{tab:teach_shape}}, and these were used to collect training data for LCFA.
For verification of the force adaptation of the learned LCFA, the workpieces shown in \tableN{\ref{tab:simpleshape}} were used in the experiments.

\begin{figure}[tp]
  \centering
    \centering
    \includegraphics[width=0.85\linewidth]{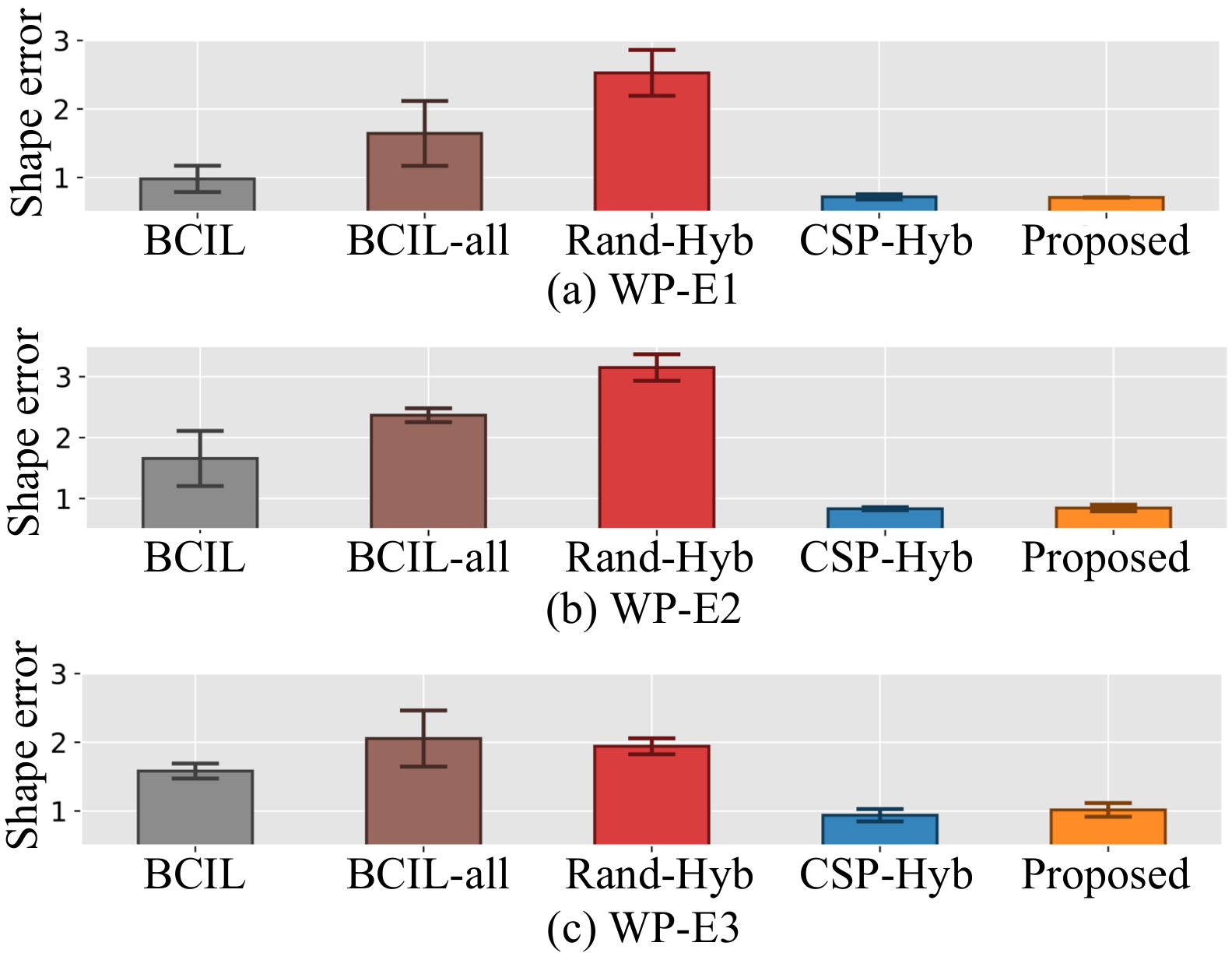}
    \caption{Comparison of shape error for evaluated methods}
    \label{fig:shape_error}
\end{figure}

\begin{figure}[tp]
  \centering
  \includegraphics[width=0.85\linewidth]{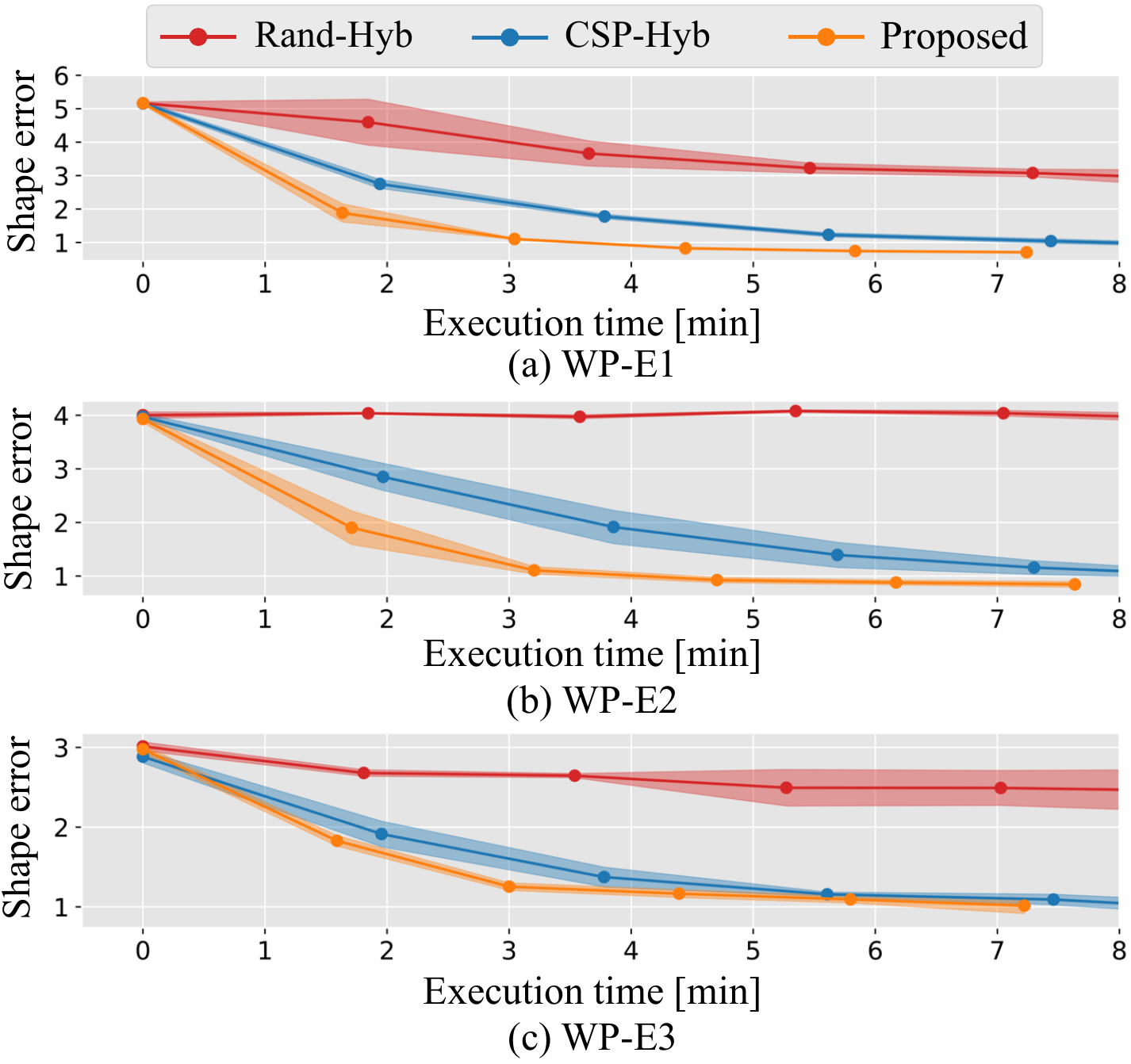}
  \caption{Execution Time and shape error for each method}
  \label{fig:result_shape}
\end{figure}

\begin{table}[tp]
    \centering
    \small
    \caption{Final shapes after grinding for each method}
    \label{tab:result_shape}
    \setlength{\tabcolsep}{2pt}
    \begin{tabular}{cccc}
      \toprule
      Name & WP-E1 & WP-E2 & WP-E3\\
      \midrule
      \raisebox{3.5\height}{BCIL}  &
      \includegraphics[width=1.8cm]{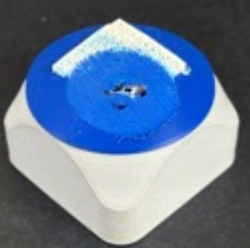} & 
      \includegraphics[width=1.8cm]{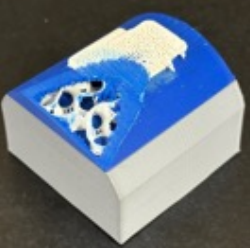} & 
      \includegraphics[width=1.8cm]{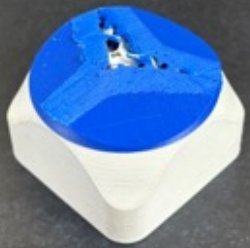} \\ 
      \midrule
      \raisebox{3.5\height}{BCIL-all}  &
      \includegraphics[width=1.8cm]{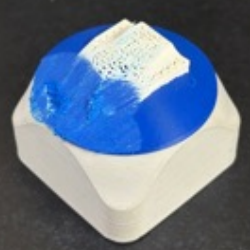} & 
      \includegraphics[width=1.8cm]{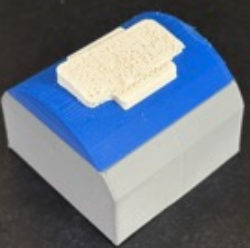} & 
      \includegraphics[width=1.8cm]{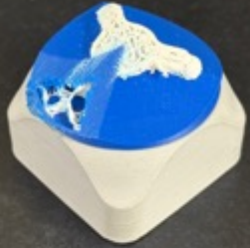} \\
      \midrule
      \raisebox{3.5\height}{Rand-Hyb}  &
      \includegraphics[width=1.8cm]{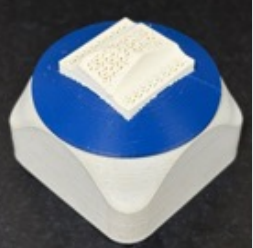} & 
      \includegraphics[width=1.8cm]{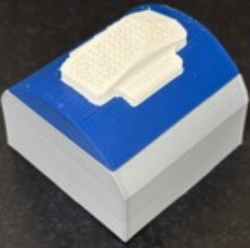} & 
      \includegraphics[width=1.8cm]{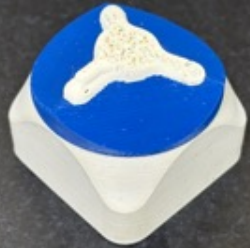} \\
      \midrule
      \raisebox{3.5\height}{CSP-Hyb}  &
      \includegraphics[width=1.8cm]{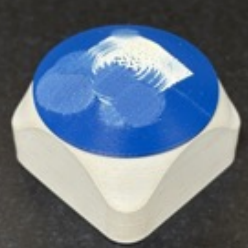} & 
      \includegraphics[width=1.8cm]{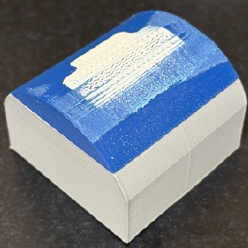} & 
      \includegraphics[width=1.8cm]{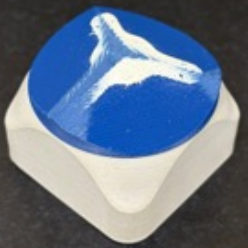} \\
      \midrule
      \raisebox{3.5\height}{Proposed}  &
      \includegraphics[width=1.8cm]{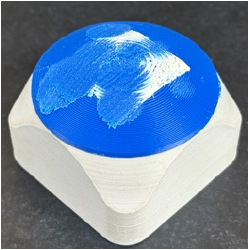} & 
      \includegraphics[width=1.8cm]{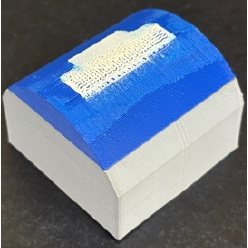} & 
      \includegraphics[width=1.8cm]{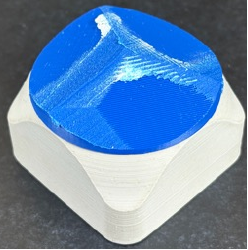} \\
      \bottomrule
      \end{tabular}

\end{table}

\subsubsection{Training of Local Contact-Force Adaptation}
Demonstrations were performed only five times for each workpiece in \tableN{\ref{tab:simpleshape}}, with each demonstration lasting approximately 6~s, for a total duration of 30 s.  
In the demonstration phase, the demonstrator adjusted the normal force so that the tangential force remained constant at 4~N.  
Training data were collected at 1000~Hz, and the trained LCFA  performed inference at 20~Hz.
The policy $\pi_w$ was trained for approximately 240 steps per episode, or 1200 steps in total.
A four-layer long short-term memory (LSTM)  \cite{lstm_hochreiter1997long} was employed to capture the time-series nature of the data, where each LSTM layer consists of 512 neurons, followed by a single fully connected layer.

All training computations were performed on a workstation equipped with an Intel Core i9-14900K CPU and an NVIDIA RTX 4090 GPU.

\subsubsection{Computation Time per Cycle}
The computational cost of each module was evaluated separately.
Shape observation and point-cloud processing required $50.5 \pm 0.5$ s on average,
most of which was spent on robot motion required to move the sensor between observation viewpoints.
Optimization of the Global Cutting-Surface Planning (GCSP) required $5.5 \pm 4.3$ s per update. The Local Contact-Force Adaptation (LCFA) policy was executed at 20 Hz during grinding, with an average inference time of $1.47 \pm 3.4$ ms per step. 
These computations were sufficiently fast to allow online execution of the proposed framework.

\section{Experimental Results}

\subsection{Evaluation of Grinding Performance with Limited training data \textnormal{(RQ1)}}
\label{subsec:resultpropose}

\subsubsection{Setup}

RQ1 investigates whether the proposed method outperforms baseline methods in grinding performance and time efficiency.
To address this question, grinding experiments were conducted on the three workpieces listed in \tableN{\ref{tab:shape}}.
The compared methods are described below.

\begin{itemize}
\item \textbf{Proposed}: The cutting surface is obtained by GCSP, and the removal shape is determined accordingly. LCFA executes the planned removal shape until the cutting surface is reached. Training data for LCFA were collected using bilateral control on the two workpieces shown in \tableN{\ref{tab:teach_shape}}.
\item \textbf{Rand-Hyb}: The cutting surface is randomly selected without using GCSP.
A hybrid controller that limits excessive force grinds the cutting surface for a fixed duration, after which a new cutting surface is randomly selected.
\item \textbf{CSP-Hyb}: The cutting surface is generated by GCSP.
A hybrid controller that limits excessive force grinds the cutting surface for a fixed duration, after which GCSP generates the next cutting surface.
\item \textbf{BCIL}: GCSP is not used, and grinding is performed using a policy learned from demonstrations of continuous grinding from start to finish. Training data were collected using bilateral control on WP-E1 in \tableN{\ref{tab:shape}}.
\item \textbf{BCIL-all}: GCSP is not used, and grinding is performed using a policy learned from demonstrations of continuous grinding from start to finish. Training data were collected using bilateral control on WP-E1, WP-E2, and WP-E3 in \tableN{\ref{tab:shape}}.
\end{itemize}

As the evaluation criterion, the shape error defined in \equationname{\ref{eq:error}} is used; however, since the precision of point-cloud observation is limited, this error does not converge to zero.
For methods using cutting surfaces, grinding is terminated after the error decreases once and the change over two consecutive updates becomes less than 5~\% of the initial shape error.
In contrast, for methods learned from continuous demonstrations without GCSP, demonstrations were performed to reach a shape error comparable to that of the cutting-surface-based method, and grinding was performed for the same duration as the demonstration.
In addition, the execution time is defined as the total duration including observation, planning, and grinding actions until the termination condition is satisfied.

\setlength{\tabcolsep}{1.5pt}
\begin{table*}[tp]
\centering
\caption{Grinding time and in-limit force ratio for WP-S by each method}
\label{tab:simple_grinding}
\begin{tabular}{ccc|cc|cc|cc|cc} 
 \toprule
  \multicolumn{1}{c}{} & \multicolumn{2}{c}{WP-S1}& \multicolumn{2}{c}{WP-S2} & \multicolumn{2}{c}{WP-S3} & \multicolumn{2}{c}{WP-S4} & \multicolumn{2}{c}{WP-S5} \\
  \cmidrule(lr){2-3}
  \cmidrule(lr){4-5}
  \cmidrule(lr){6-7}
  \cmidrule(lr){8-9}
  \cmidrule(lr){10-11}
  \multicolumn{1}{c}{Method} & \makecell{Execution\\Time [s]} & \makecell{In-limit\\force ratio [\%]} & \makecell{Execution\\Time [s]} & \makecell{In-limit\\force ratio [\%]} & \makecell{Execution\\Time [s]} & \makecell{In-limit\\force ratio [\%]} & \makecell{Execution\\Time [s]} & \makecell{In-limit\\force ratio [\%]} & \makecell{Execution\\Time [s]} & \makecell{In-limit\\force ratio [\%]} \\ 
 \midrule
  Demo-Speed-1        &$5.3  \pm 0.1$&$100$&$-$           &$  0$&$-$           &$  0$&$-$           &$  0$&$-$           &$   0$\\
  Demo-Speed-2        &$53.2 \pm 0.2$&$100$&$40.5 \pm 0.3$&$100$&$40.2 \pm 0.6$&$100$&$27.6 \pm 0.1$&$100$&$27.5 \pm 0.1$&$100 $\\ 
  Proposed            &$5.3  \pm 0.1$&$100$&$\mathbf{6.0  \pm 0.2}$&$100$&$\mathbf{9.2  \pm 0.2}$&$100$&$\mathbf{10.9 \pm 0.3}$&$100$&$27.8 \pm 2.5$&$100$\\
\bottomrule 
 \end{tabular}
\end{table*}

\subsubsection{Results}
Figure~\ref{fig:shape_error} compares the shape errors,
Table~\ref{tab:result_shape} illustrates the resulting shapes,
and \figureN{\ref{fig:result_shape}} plots the execution time
and the evolution of the shape error.
From \figureN{\ref{fig:shape_error}}, BCIL and BCIL-all without GCSP fail to sufficiently reduce the final shape error.
As shown in \tableN{\ref{tab:result_shape}}, excessive removal beyond the target shape is observed.
Even when the removal shape is determined based on cutting surfaces, the shape error is not reduced if the cutting surfaces are not properly planned, as in \textbf{Rand-Hyb}.
In contrast, \textbf{Proposed} and \textbf{CSP-Hyb} sufficiently reduce the shape error by using GCSP.
From \figureN{\ref{fig:result_shape}}, for shapes such as WP-E3 with small contact area and low density, the execution time of \textbf{Proposed} and \textbf{CSP-Hyb} are comparable.
In contrast, for shapes such as WP-E1 and WP-E2 with large contact area and high density, \textbf{Proposed} completes grinding in a shorter time than \textbf{CSP-Hyb}.

These results indicate that our proposed method, DecompGrind, achieves time-efficient grinding for workpieces with different shapes and material hardness.

\begin{figure}[tp]
\centering
\includegraphics[width=0.85\linewidth]{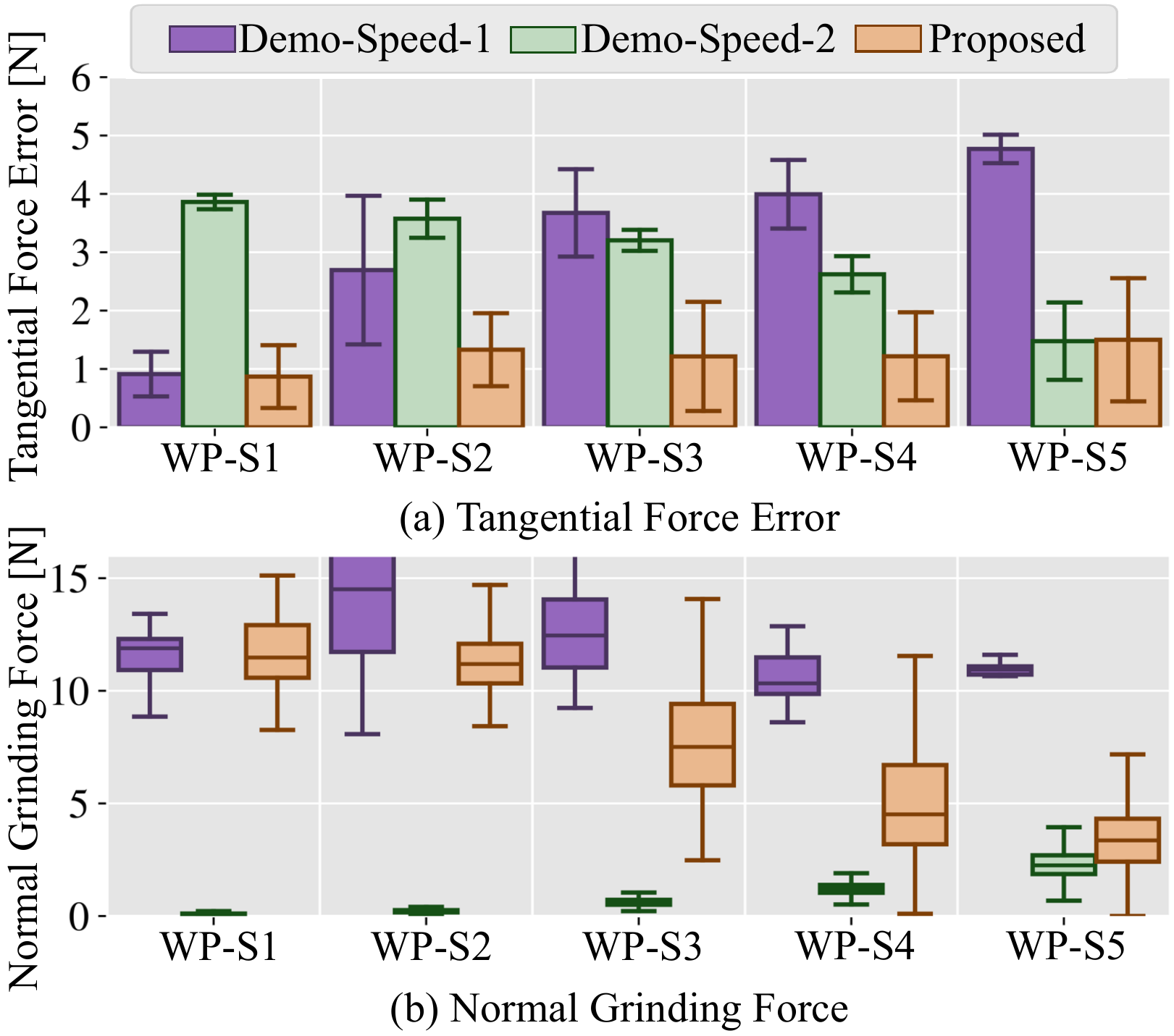}
\caption{Comparison of grinding forces for WP-S between proposed method and methods without LCFA.
Force data for $t \geq 3~s$ are shown.
(a) Tangential force error along the $\mathrm{Z}$-axis relative to the target tangential force of 4~N.
(b) Normal force along the $\mathrm{X}$-axis.
}
\label{fig:force}
\end{figure}

\subsection{Force Adaptation Behavior of Learned Policy \textnormal{(RQ2)}}
\label{subsec:simpleshape}

\subsubsection{Setup}

RQ2 investigates whether the learned LCFA  adjusts the force appropriately in response to removal resistance caused by differences in shape and material hardness.
Simple workpieces that produce relatively uniform removal resistance are used to evaluate the time efficiency of grinding.
The in-limit force ratio is defined as the ratio of grinding within the allowable tangential force limit of 9~N.
The workpieces used in the evaluation are shown in \tableN{\ref{tab:simpleshape}}, and the comparison methods are described below.

\begin{itemize}
\item \textbf{Demo-Speed-1}: Grinding speed in the normal direction ($\mathrm{X}$-axis) is set to a constant value matching the demonstration speed on WP-T1 in \tableN{\ref{tab:teach_shape}}.
\item \textbf{Demo-Speed-2}: Grinding speed in the normal direction ($\mathrm{X}$-axis) is set to a constant value matching the demonstration speed on WP-T2 in \tableN{\ref{tab:teach_shape}}.
\end{itemize}

\subsubsection{Results}

The tangential force error and the normal force for the simple workpieces are shown in Fig.~\ref{fig:force}, and the grinding time and the ratio of grinding within the allowable force are summarized in \tableN{\ref{tab:simple_grinding}}.
Fig.~\ref{fig:force}(a) shows the error with respect to the target tangential force (4 N).
\textbf{Proposed} maintains consistently small errors with respect to the target tangential force across different workpieces.
In contrast, \textbf{Demo-Speed-1} and \textbf{Demo-Speed-2} achieve errors comparable to \textbf{Proposed} only when the workpiece used for speed tuning matches the test workpiece; otherwise, the error increases.

From \figureN{\ref{fig:force}}(b), the normal force in the proposed method decreases for shapes with larger contact areas and higher density.
In \textbf{Demo-Speed-1}, the pressing force is not sufficiently adapted to differences in shape and density, and excessive forces appear for some shapes.
In \textbf{Demo-Speed-2}, insufficient pressing force is observed for certain workpieces.
Furthermore, \tableN{\ref{tab:simple_grinding}} shows that \textbf{Proposed} completes grinding in a shorter time while keeping the tangential force within the allowable range (below 9~N).

These results indicate that LCFA improves time efficiency by adapting the contact force.

\begin{table*}[th]
\centering
\caption{Execution time and in-limit force ratio for WP by each method}
\label{tab:all_Execution_time}
\begin{tabular}{ccc|cc|cc} 
 \toprule
 \multicolumn{1}{c}{} & \multicolumn{2}{c}{WP-E1}& \multicolumn{2}{c}{WP-E2} & \multicolumn{2}{c}{WP-E3} \\ 
  \cmidrule(lr){2-3}
  \cmidrule(lr){4-5}
  \cmidrule(lr){6-7}
  \multicolumn{1}{c}{Method} & Execution Time [s] & \makecell{In-limit\\force ratio [\%]} & Execution Time [s] & \makecell{In-limit\\force ratio [\%]} &  Execution Time [s] & \makecell{In-limit\\force ratio [\%]}\\ 
 \midrule
Demo-Speed-1    &$405.7 \pm 4.3$ & $20 $&$569.4 \pm 7.0$&$16$&$\mathbf{178.7 \pm 0.9}$&$100$\\
Demo-Speed-2    &$243.8 \pm 3.3$ & $\mathbf{100} $&$212.6 \pm 1.7$&$\mathbf{100} $&$239.6 \pm 0.6$&$100$\\ 
Proposed             &$\mathbf{182.6 \pm 0.7}$ & $\mathbf{100} $&$\mathbf{192.3 \pm 3.1}$&$\mathbf{100}$&$\mathbf{180.0 \pm 1.7}$&$100$\\
\bottomrule 
 \end{tabular}
\end{table*}

\begin{table}[t]
\centering
\caption{Comparison of execution time and in-limit force ratio with different demonstration workpieces}
\label{tab:com_demo}
\begin{tabular}{ccccccc} 
 \toprule
 \multicolumn{1}{c}{} & \multicolumn{2}{c}{WP-E2}  \\ 
  \cmidrule(lr){2-3}
  \multicolumn{1}{c}{Method} & Execution Time [s] & \makecell{In-limit\\force ratio [\%]}\\ 
 \midrule
 Training WP-E             &$259.3 \pm 10.3$ & $60 $\\ 
 Training WP-T             &$\mathbf{192.3 \pm 3.1}$ & $\mathbf{100}$\\
\bottomrule 
 \end{tabular}
\end{table}

\subsection{Efficiency Improvement from Learning-Based Force Adaptation \textnormal{(RQ3)}}
\label{subsec:resultfai}

\subsubsection{Setup}
RQ3 evaluates the improvement in grinding efficiency achieved by force adaptation by comparison with methods that do not adapt the contact force.
The same GCSP is used across all methods to plan the removal shape, and the comparison methods are identical to those in \textnormal{RQ2}.

As performance metrics, we evaluate the total grinding time required to reach a predefined target shape error threshold and whether the removal resistance force is properly regulated.
The effect of force adaptation is prominent during large-volume removal but becomes limited near the target shape.
Accordingly, the target shape error is defined as 20~\% of the difference between the initial shape error and the final shape error reported in \textnormal{RQ1}.

\subsubsection{Results}
\tableN{\ref{tab:all_Execution_time}} shows a comparison of grinding execution times.
From \tableN{\ref{tab:all_Execution_time}}, in \textbf{Demo-Speed-1}, grinding stops for WP-E1 and WP-E2 due to excessive force beyond the allowable limit, resulting in a longer execution time.
In contrast, both \textbf{Demo-Speed-2} and \textbf{Proposed} complete grinding within the force limit, and \textbf{Proposed} removes material in a shorter time.

These results indicate that the proposed method outputs pressing force according to removal resistance and achieves stable grinding in a shorter time without generating excessive force, even for workpieces with different shapes and densities.

\subsection{Effect of Workpieces Used for LCFA Training \textnormal{(RQ4)}}
\label{subsec:resultteachingshape}

\subsubsection{Setup}
In the proposed method, demonstration data are collected using the simple workpieces shown in \tableN{\ref{tab:teach_shape}}, which are designed to facilitate force adaptation.
To investigate RQ4, we compare policies trained using demonstration data collected from different training workpieces, namely those in \tableN{\ref{tab:simpleshape}} and \tableN{\ref{tab:shape}}.
The comparison methods are defined as follows.

\begin{itemize}
  \item \textbf{Training WP-T}: Training data are collected using the two simple workpieces in \tableN{\ref{tab:teach_shape}}, and the policy is trained using these data.
  \item \textbf{Training WP-E}: Training data are collected using WP-E1 in \tableN{\ref{tab:shape}} for removal shapes segmented by GCSP, and the policy is trained accordingly. Total number of training steps is adjusted to be comparable to that of \textbf{Training WP-T}.
\end{itemize}

For performance evaluation, experiments were conducted on the WP-E2.

\subsubsection{Results}
The results of LCFA trained with different workpieces for training data collection are shown in \tableN{\ref{tab:com_demo}}.
From \tableN{\ref{tab:com_demo}}, \textbf{WP-T} achieves shorter grinding time than \textbf{WP-E}.
In addition, \textbf{WP-E} infers excessively large forces, causing the grinding process to stop,
whereas \textbf{WP-T} completes material removal while remaining within the allowable force limit.

These results suggest that providing demonstrations covering the minimum and maximum contact areas and densities appearing in the target workpieces permits appropriate force adaptation to be learned with a small amount of data.

\section{DISCUSSION}
The experimental results suggest that decomposing the grinding process into geometric removal-shape planning and local contact-force adaptation is effective for robotic grinding. GCSP determines a stable global removal sequence toward the target shape, while LCFA adapts the contact force to variations in removal resistance caused by differences in contact area and material density. This separation of tasks restricts learning to local contact-force adaptation and allows the policy to be learned from a limited number of demonstrations.

In this study, the evaluation focused on workpieces with different shapes and densities under fixed belt conditions. 
When substantially different belt characteristics or material hardness are introduced, the resulting force scale may change beyond the range considered in this work. 
Extending our framework to accommodate a wider force range may require appropriate scaling strategies, including adjustment of the bilateral control gains \cite{bi_gain_LI2023103057}.

One limitation of the proposed method is that grinding follows straight trajectories, making continuous removal along curved surfaces such as cylindrical faces difficult and thus reducing grinding efficiency.
Although curved trajectories can be considered, the motion and force directions vary with surface curvature, complicating force demonstration and learning.
In removal processes using a grinding belt, the dominant removal resistance appears mainly along the belt normal and moving directions.
Since the proposed force adaptation is defined based on these components, the same rule could be applied even when posture or contact conditions change.
Therefore, straight motion is used as the basic trajectory, while the robot posture is continuously varied near the target shape according to surface curvature.
If such posture variation could be planned using diffusion models  \cite{hachimine2024cutting} or target-shape-based path generation methods  \cite{cube_tafuro2024autonomous}, the proposed method could be extended to curved-surface grinding without needing to modify the force demonstration framework.

\section{CONCLUSION}
We proposed DecompGrind, a framework for time-efficient robotic grinding under conditions of varying workpiece shapes and material properties. The proposed method decomposes the grinding task into global removal-shape planning and local contact-force adaptation based on contact feedback. By separating geometric shape planning from learning-based force adaptation, the proposed framework restricts the scope of the learning problem in robotic grinding. As a result, force adaptation can be learned from a small amount of demonstration data. Experimental results using real robotic systems demonstrate that the proposed method achieves time-efficient grinding of workpieces having different shapes and densities while also maintaining safe contact forces.


\bibliographystyle{IEEEtran}
\bibliography{reference}
\end{document}